\documentclass{article}
\usepackage[final,main]{neurips_2026}
\usepackage[utf8]{inputenc}
\usepackage[T1]{fontenc}
\usepackage{hyperref}
\usepackage{url}
\usepackage{booktabs}
\usepackage{amsfonts}
\usepackage{amsmath}
\usepackage{nicefrac}
\usepackage{microtype}
\usepackage{graphicx}
\usepackage{xcolor}
\usepackage{enumitem}
\usepackage{multirow}
\usepackage{subcaption}
\usepackage{wrapfig}
\usepackage{needspace}

\title{Transferable, Time-Parallel Graph Dynamics\\via Space-Time Factorization}

\author{Wencong Yang \quad Chaopeng Shen\thanks{Corresponding author: \texttt{cxs1024@psu.edu}.} \quad Abdolmehdi Behroozi\\
Department of Civil and Environmental Engineering\\
Penn State University\\
University Park, PA 16802-1408, USA\\
\texttt{wby5078@psu.edu}, \texttt{cxs1024@psu.edu}, \texttt{amb10399@psu.edu}}

\begin{document}

\maketitle

\begin{abstract}
Time-parallel neural operators (FNO and successors) have transformed PDE surrogates on \emph{regular} grids. Mesh-based operators (MeshGraphNets, Transolver) extend operator learning to \emph{irregular} geometries, while jointly learning across diverse topologies and retaining non-recurrent temporal evaluation remains challenging. We propose \textbf{space-time factorization}: partition the model into local temporal operators (learned, graph-independent) and a spatial mixer that propagates information across nodes without graph-dependent learned parameters. This structural condition allows the same parameters to run on unseen graphs; non-recurrent instances also admit time-parallel evaluation. Accurate zero-shot transfer further depends on controlling three information leaks---graph-dependent weights, graph-specific features, and graph-correlated training distributions---and on suitable task physics and training coverage. The same protocol applies to two distinct spatial-mixer families tested here---fixed-spectral and learned-attention---suggesting the principle is not tied to a single mixer choice. We validate the principle in three domains. On simulating floodwave propagation in river networks, a 27K-parameter model trained on the Yangtze transfers zero-shot to the Mississippi with median NSE $0.998$ in the boundary-conditioned operational evaluation, training $5.5\times$ faster than a stabilized recurrent baseline in a controlled one-epoch workload. On cross-geometry Navier--Stokes, the three-seed SepFNO mean relative vorticity error on five unseen targets falls by about 28\% from 5 to 12 training geometries, with a complementary single-run SepTransolver trend. On cross-city traffic forecasting, zero-shot transfer reaches within $12.9\%$ of an oracle trained on the target city. A consistent pattern emerges across these settings: cross-geometry error decreases with both the \emph{number} of training geometries and their \emph{relatedness} to the test, with the rate of accumulation depending on the spatial mixer. The result opens the path for pooled-data, foundation-style models for graph dynamics on irregular topologies.
\end{abstract}

\section{Introduction}

Operator learning on regular grids has become one of the dominant paradigms for fast Partial Differential Equations (PDE) surrogates: time-parallel models such as Fourier Neural Operators (FNO) report up to three orders-of-magnitude faster solves than traditional solvers~\citep{li2021fourier}, and avoiding recurrent rollout also avoids the substantial activation-storage burden of backpropagation through time on long sequences~\citep{gruslys2016memory}. Extending these benefits to irregular geometries---and, more broadly, to spatiotemporal prediction on graphs that arise in physical and engineered systems, including flood propagation in river networks, fluid dynamics on unstructured meshes, and traffic forecasting on road networks---is an open problem. The workhorse architectures for these tasks---graph neural operators, MeshGraphNets, and spatiotemporal graph neural networks (GNNs)---couple message passing with recurrent time stepping: at each time step, a graph operator propagates information across nodes, and a Gated Recurrent Unit (GRU) or LSTM updates node states. This \emph{entangled} design has an underappreciated consequence: it precludes access to time-parallel solvers. When a nonlinear graph operation sits inside the temporal recurrence (e.g., $h_{t+1} = \sigma(W A h_t + \cdots)$), the per-node temporal evolution is no longer linear time-invariant, the system can only be advanced sequentially, and standard tools from scientific computing---Fast Fourier Transform (FFT), spectral kernels, classical time-parallel integration algorithms (see~\S\ref{sec:related})---are inapplicable. It is even unclear whether there is a viable path for using time-parallel algorithms to simulate complex nonlinear graph dynamics such as Navier--Stokes, since classical Parareal and MGRIT converge poorly on advection-dominated regimes. The same entanglement also degrades cross-graph transfer: even with shared weights~\citep{hamilton2017inductive,gilmer2017mpnn}, the learned dynamics become implicitly conditioned on training-graph topology when information leakage are present, limiting reliable transfer and pooled training across graphs.

We propose \textbf{space-time factorization} as the structural condition that removes both obstructions: partition the architecture into (a)~\emph{local temporal operators} whose learned parameters are graph-independent and act per node, and (b)~\emph{spatial mixers} that propagate information across nodes through fixed graph operators with no learned parameters or adjacency-free attention with shared learned projections. Two consequences follow: \textbf{efficiency} (parallel temporal solve without backpropagation through time, spatial mix in $O(K|E|)$ where $K$ is the number of propagation steps and $|E|$ the number of graph edges) and \textbf{cross-graph transfer and knowledge accumulation} (no learned parameter touches the graph, so models deploy zero-shot by swapping the adjacency matrix, and data pooled from many topologies trains a single model whose accuracy can improve with broader training-geometry coverage, with gains depending on pool composition). The efficiency benefit is direct: once temporal and spatial operators are decoupled, the temporal step becomes per-node and linear-in-time, opening access to FFT, spectral kernels, time-parallel attention, and other non-recurrent solvers.

Our contributions: (1)~\textbf{space-time factorization} as a design principle for graph dynamical systems, where local temporal operators are paired with fixed graph propagation; (2)~a \textbf{systematic analysis} of three information leaks that break the factorization---graph-dependent weights, graph-dependent features, and graph-correlated training distributions---and the \emph{time parallelism} and empirical pattern of \emph{knowledge accumulation across training geometries} that the principle enables; (3)~\textbf{validation} across three spatiotemporal domains plus a static one, with two negative controls that map out the applicability boundary.

\section{Related Work}
\label{sec:related}

\textbf{Neural operators on regular grids.}
The Fourier Neural Operator~\citep{li2021fourier} stacks FFT-based spectral kernels and predicts an entire output window in a single forward pass: the temporal solve is $O(T \log T)$, BPTT is avoided, and hundreds of time steps cost nearly the same as one~\citep{gruslys2016memory}. Standard FNO assumes a regular grid; extending the same time-parallel benefits to irregular meshes is the open problem we target.

\textbf{Operator learning on irregular geometries.}
Mesh-based and geometry-aware operator models---MeshGraphNets~\citep{pfaff2021meshgraphnets}, Geo-FNO~\citep{li2023geofno}, GINO~\citep{li2023gino}, AMG~\citep{li2025amg}, RIGNO~\citep{mousavi2025rigno}, Sp$^2$GNO~\citep{sarkar2025sp2gno}, and Transolver~\citep{wu2024transolver}---extend operator learning to irregular domains, but two patterns recur: spatial operators carry mesh-dependent learned weights (relative-position edges, geometry-conditioned encoders) and time advances autoregressively, consuming substantial compute and memory on long horizons. Our work imposes graph-independent learned weights and a time-parallel temporal solver simultaneously; Appendix~\ref{sec:related_benchmarks} compares cross-geometry benchmarks (PDEBench, SURF, MGN cross-geometry studies).

\textbf{Classical time-parallel methods.}
Parareal~\citep{lions2001parareal,gander2007parareal} and MGRIT~\citep{falgout2014mgrit} have well-established convergence on linear, diffusion-dominated PDEs but converge slowly and non-robustly on nonlinear advection-dominated problems including Navier--Stokes~\citep{desterck2020paraconv}. Operator-splitting (Chorin--Temam) decouples partially but still advances time sequentially. FNO sidesteps both by \emph{learning} per-layer-linear spectral kernels stacked with nonlinear activations---on regular grids. Whether this trade-off transfers to irregular meshes with graph-free spatial mixing is the empirical question our PDE experiments answer.

\textbf{Spatiotemporal graph neural networks.}
DCRNN~\citep{li2018dcrnn}, STGCN~\citep{yu2018stgcn}, Graph WaveNet~\citep{wu2019graphwavenet}, and AGCRN~\citep{bai2020agcrn} all interleave learned dynamics with the graph: STGCN composes learned per-edge weights with temporal blocks; Graph WaveNet learns adaptive adjacency; DCRNN places diffusion \emph{inside} the GRU update; AGCRN ties node-specific embeddings into the recurrence. These choices affect temporal parallelism and transfer in different ways: recurrence limits time parallelism, while source-node-specific embeddings directly obstruct transfer. Our SepGRU keeps the GRU temporal core (appropriate for the short prediction horizon of standard traffic benchmarks) but moves graph propagation outside the recurrence with feature-conditioned, graph-free hop weights---demonstrating that the factorization principle applies even when FFT-in-time is impractical.

\textbf{Inductive and decoupled graph learning.}
Inductive methods---MPNNs~\citep{gilmer2017mpnn}, GraphSAGE~\citep{hamilton2017inductive}, GCNs~\citep{kipf2017semi}---share parameters across nodes and graphs but are evaluated almost exclusively on static problems. SGC~\citep{wu2019simplifying}, SIGN~\citep{frasca2020sign}, and APPNP~\citep{klicpera2019predict} decouple fixed propagation from graph-free predictors, with graphon analyses studying filter transferability~\citep{ruiz2021graph}. Extending this to nonlinear temporal dynamics is non-trivial: temporal evolution opens leakage channels (magnitude distributions, boundary-feature memorization) that have no analogue in static node classification, and time-parallelism becomes available only when per-step linearity is enforced layer-by-layer.

\section{Method: Space-Time Factorization}
\label{sec:method}

\subsection{Problem Setting}

Consider a spatiotemporal prediction task on a graph $G = (V, E)$ with adjacency matrix $\mathbf{A} \in \mathbb{R}^{N \times N}$. Each node $i$ has local features $\mathbf{x}_i \in \mathbb{R}^d$ and a time-varying target $\mathbf{y}_i(t)$. We want a model $f_\theta$ trained on graphs $\{G_1, \ldots, G_K\}$ that generalizes to an unseen graph $G_\text{test}$ with different $N$, different $\mathbf{A}$, but the same type of node features, dynamics, and prediction task.

\subsection{The Factorization Principle}

Standard spatiotemporal GNNs interleave learned message passing with recurrent time stepping: at each time $t$, they compute hidden states $\mathbf{h}(t) = \text{GRU}(\mathbf{h}(t{-}1),\; \text{MP}(\mathbf{h}(t{-}1), \mathbf{A}))$, where GRU is a Gated Recurrent Unit~\citep{cho2014gru} (any recurrent cell with learned dynamics works equally well for our argument), MP is graph message passing, and $\mathbf{h}(t) \in \mathbb{R}^{N \times d}$ is the snapshot of $d$-dimensional hidden states across all $N$ nodes at time $t$ (not a time series). This update couples learned parameters $\theta$ with the adjacency $\mathbf{A}$ at every step. We propose factorizing these:

\begin{quote}
\textbf{Space-Time Factorization:} Partition the model into (a)~\emph{local temporal operators} with learned parameters $\theta$ acting on each node independently, and (b)~a \emph{spatial mixer} $\mathcal{M}$ that propagates information across nodes using \emph{no graph-dependent learned parameters}. Temporal and spatial operations compose but do not interleave.
\end{quote}

\begin{figure}[!htbp]
\centering
\includegraphics[width=\textwidth]{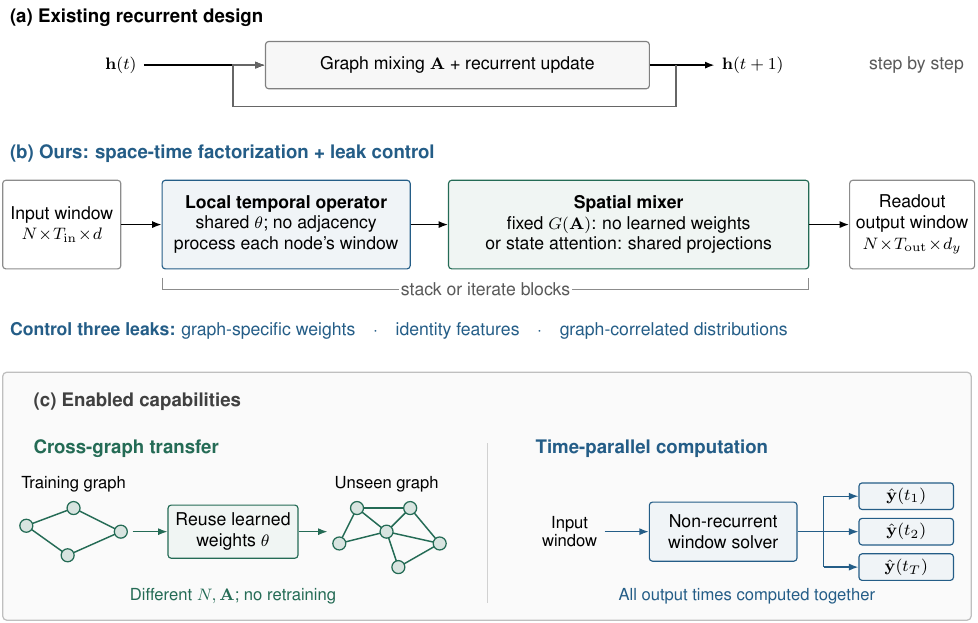}
\caption{Method schematic: space-time factorization and its enabled capabilities. $N$ is the node count, $\mathbf A$ the adjacency matrix, $\theta$ the shared learned parameters, and $T_{\rm in}/T_{\rm out}$ the input/output window lengths.}
\label{fig:method_schematic}
\end{figure}

Figure~\ref{fig:method_schematic} contrasts graph mixing inside a recurrent update with our factorized window-based template.

Formally, the forward pass has the structure:
\begin{equation}
    \mathbf{h}_i(t) = f_\theta\!\left(\mathbf{x}_i(t),\; \big[\mathcal{M}(\mathbf{h}(t); \mathbf{A})\big]_i\right)
    \label{eq:separation}
\end{equation}
where the spatial mixer $\mathcal{M}$ propagates information across nodes \emph{without learned parameters that condition on} $\mathbf{A}$. This admits two construction patterns: (a) a \textbf{fixed graph operator} $\mathcal{M}(\mathbf{h}; \mathbf{A}) = G(\mathbf{A})\mathbf{h}$ where $G(\mathbf{A}) \in \{\mathbf{A}, \mathbf{A}^k, \sum_k \alpha_k \mathbf{A}^k, T_k(\mathbf{A})\}$ denotes fixed one-hop propagation, $k$-hop propagation, weighted multi-hop diffusion, or Chebyshev polynomial graph filtering, respectively, with no graph-specific learned parameters (Our SepFNO, SepGFNO, SepGRU below); the rich local-feature information enters via $\mathbf{h}$, while $G(\mathbf{A})$ propagates those feature-encoded states through the fixed adjacency. Or (b) a \textbf{learned cross-node operator} (e.g., self-attention or Physics-Attention slice tokens in our SepTransolver below) whose learned weights ($Q, K, V$ projections, slice projection $W$) act on \emph{node states} but never reference $\mathbf{A}$---node-to-node coupling emerges from softmax similarities of the states themselves. Both patterns satisfy the constraint that no learned parameter encodes graph identity (in pattern (b), the mesh may still be used during preprocessing to compute features such as distance-to-boundary, but is never consulted inside any layer). $f_\theta$ is a shared temporal operator (MLP, FFT kernel, or GRU) applied identically to every node.

\paragraph{Iterating for spatial reach and nonlinearity.}
Equation~\eqref{eq:separation} shows a single application of $\mathcal{M}$; practical instantiations stack or iterate this operation. For fixed graph operators, stacking can be done by repeated propagation, higher-order graph polynomials, or multiple operator layers. For learned graph-free cross-node operators, multiple attention layers play can be stacked. What the factorization forbids is not iteration but \emph{graph-conditioned entanglement}: spatial mixing must not be interleaved with learned parameters that depend on $\mathbf{A}$. The composition $f_\theta \circ \mathcal{M} \circ f_\theta \circ \mathcal{M} \circ \cdots$ preserves graph-independence layer-by-layer while recovering nonlinearity through stacking. This extends the SGC~\citep{wu2019simplifying} / SIGN~\citep{frasca2020sign} decoupling logic from static classification to spatiotemporal prediction, enabling efficiency and knowledge-accumulation benefits at the cost of additional leak control (next subsection).

\subsection{Information Leaks: Three Levels of Graph Dependence}
\label{sec:leak}

Standard inductive GNNs already share weights across nodes, but graph-specific information can still leak into the learned representation through weights, features, and training distributions. We therefore impose three design constraints.

\paragraph{Level 1: Graph-free weights.}
All learned parameters must be independent of the graph. Per-node learnable embeddings, Laplacian positional encodings~\citep{dwivedi2020lpe}, and learned adjacency matrices violate this; SGC~\citep{wu2019simplifying}, SIGN~\citep{frasca2020sign}, GraphSAGE~\citep{hamilton2017inductive}, and our SepFNO/SepGRU/SepGFNO satisfy it.

\paragraph{Level 2: Graph-free features.}
Input features must describe transferable local properties rather than graph-specific identity. Absolute or normalized coordinates are common leaks; local descriptors with consistent meaning across graphs (distance to boundary, curvature, slope) are acceptable.

\paragraph{Level 3: Graph-correlated training distributions.}
Even with graph-free weights and features, the training distribution can carry graph-specific signatures (e.g., characteristic magnitude ranges per geometry); a model can then infer graph identity from data statistics. Distribution augmentation---rotation, magnitude scaling, feature dropout---is the standard remedy for breaking these correlations. The remedy is architecture-dependent: magnitude perturbation helps fixed-polynomial mixers but can hurt softmax-based attention whose weights are activation-scale sensitive (\S\ref{sec:pde}).

\subsection{Per-Step Linearity as the Enabler of Time-Parallelism}
\label{sec:efficiency}

We sidestep classical Parareal/MGRIT difficulties (\S\ref{sec:related}) by following the FNO trade-off: each temporal kernel is FFT-able through per-step linearity, while stacking with nonlinear activations recovers nonlinear input--output behavior. Our SepTransolver instead achieves time-parallelism through windowed channel I/O (non-recurrent single forward pass, $O(T)$ scaling vs SepFNO's $O(T\log T)$). Strict factorization cannot model state-dependent spatial coupling without more layers or hybrid extensions; the tradeoff is favorable when coupling is primarily local-attribute-dependent.

\subsection{Necessary Conditions for Applicability}
\label{sec:applicability}

Because no learned parameter may depend on a particular graph, two task conditions matter: (i) connectivity can be separated from the governing physics (when edge weights themselves carry equation-defining coefficients, as with the admittance matrix in power flow, this separation fails); and (ii) features carry local physics in transferable form---reach attributes for rivers, boundary descriptors for irregular-mesh PDEs, and sensor statistics for traffic. These conditions do not guarantee accuracy; leak control (\S\ref{sec:leak}), training coverage, and backbone capacity also matter. A non-recurrent temporal solver is an additional requirement for time parallelism (\S\ref{sec:efficiency}). The negative controls of \S\ref{sec:negative} test conditions (i) and (ii): power flow violates (i), airport networks violate (ii).

\subsection{Instantiations Across Domains}
\label{sec:instantiations}

All instantiations share a template: a shared encoder maps local attributes to states, graph-independent heads produce coefficients, a shared temporal processor evolves each node, and a spatial mixer couples nodes. Their temporal solvers and spatial mixers differ (Table~\ref{tab:instantiations_main}; further details in Appendix~\ref{sec:architectures_appendix}). SepFNO uses fixed Chebyshev polynomials, while SepTransolver uses adjacency-free learned Physics-Attention within this template.

\begin{table}[t]
\centering
\caption{Shared-parameter instantiations. Cross-graph deployment is defined for all five; time parallelism applies only to the non-recurrent dynamics.}
\label{tab:instantiations_main}
\small
\setlength{\tabcolsep}{4pt}
\begin{tabular}{@{}llll@{}}
\toprule
Model & Shared temporal component & Spatial mixer & Time-parallel? \\
\midrule
SepGFNO & MC coefficients, FFT-IIR & fixed upstream propagation & Yes \\
SepFNO & temporal FFT kernels & fixed Chebyshev polynomials & Yes \\
SepTransolver & windowed input/output & learned state attention & Yes \\
SepGRU & shared GRU & fixed adjacency powers & No \\
SepHybrid & static MLP & fixed pre-smoothing & N/A (static) \\
\bottomrule
\end{tabular}
\end{table}

\paragraph{River routing (SepGFNO, known physics).}
The task estimates the propagation of floodwaves downstream across thousands of time steps through a river network---a critical procedure for issuing flood warnings. The Muskingum--Cunge (MC) method solves a simplified form of the Saint-Venant continuity and momentum equations over large open-channel networks across many time steps. A per-reach MLP maps local attributes (slope, roughness, upstream area) to MC coefficients of the linear recurrence~\citep{bindas2024dmc}: $Q_i(t) = c_1 I_i(t) + c_2 I_i(t{-}1) + c_3 Q_i(t{-}1) + c_4 q'_i(t)$. Spatial propagation iterates the fixed upstream-connectivity matrix $\mathbf{N}$ for $D$ rounds (network diameter; Appendix~\ref{sec:sepgfno_detail}).

\paragraph{PDE solving (SepFNO and SepTransolver).}
The task is solving 2D incompressible Navier--Stokes (vorticity formulation) on irregular meshes---the classical operator-learning benchmark---and zero-shot transferring to a held-out geometry. Each mesh node carries local boundary features (distance to boundary, surface normal, curvature) and \emph{no absolute coordinates}. SepFNO applies per-frequency FNO-style spectral kernels and mixes neighboring information via fixed Chebyshev polynomials $T_0(\mathbf{A}), \ldots, T_K(\mathbf{A})$. SepTransolver replaces the Chebyshev mixer with Physics-Attention slice tokens (learned weights that do not reference $\mathbf{A}$) and uses windowed channel I/O instead of FFT. Both are time-parallel under our protocol (Appendices~\ref{app:sepfno}, \ref{sec:transolver_aug_appendix}).

\paragraph{Traffic forecasting (SepGRU, learned dynamics).}
The task is short-horizon traffic-speed forecasting on road sensor networks (the standard DCRNN benchmark) and zero-shot transferring across cities with disjoint sensor IDs. Each sensor has local statistics (mean speed, variance, congestion fraction). A shared GRU encodes each sensor's series independently; an MLP maps local statistics to per-node hop weights $\boldsymbol{\alpha}_i$ for fixed multi-hop diffusion $\sum_k \alpha_{i,k} (\mathbf{A}^k \mathbf{h})_i$. The graph enters only through precomputed adjacency powers (Appendix~\ref{app:sepgru}).

\paragraph{Citation classification (SepHybrid, static graphs, breadth check).} SIGN-style pre-smoothing followed by graph-free MLP heads (Appendix~\ref{sec:citation_appendix}) to check if our principle extends to static graphs.

\section{Experiments}
\label{sec:experiments}

We evaluate whether graph-independent learned components, fixed graph operators, and leakage control improve zero-shot transfer across structurally different graphs. The experiments cover the three spatiotemporal domains and the static breadth-check domain introduced in \S\ref{sec:instantiations}, plus two negative controls. In domains where the temporal dynamics admit convolutional or physics-structured solvers, the same separation also provides computational advantages.

\subsection{River Flood Routing}
\label{sec:river}

A single SepGFNO model trained on bounded-diameter Yangtze subgraphs ($D_{\max}=50$, 50{,}859 reaches) is deployed zero-shot on the Mississippi basin with no basin-specific adaptation, testing whether a physics-anchored, locally parameterized routing model transfers across a substantially different network while remaining scalable to basin-wide inference. The target is simulated reference discharge with known lateral inflows and reach attributes.

\begin{table}[ht]
\centering
\caption{Yangtze$\to$Mississippi operational evaluation on the common 86{,}133-reach cohort. Peak error is absolute percent; timing is absolute hours. Boundary flows at trunk injections use reference discharge; metrics cover 2{,}400 hourly steps per reach, not a calendar year.}
\label{tab:river}
\small
\setlength{\tabcolsep}{5pt}
\begin{tabular}{@{}lccccc@{}}
\toprule
Model & NSE med. & KGE med. & Peak (\%) & Timing (h) & NSE P10 \\
\midrule
SepGFNO, trained & \textbf{0.998} & \textbf{0.983} & \textbf{1.51} & \textbf{1} & \textbf{0.985} \\
SepGFNO, anchor-only & 0.569 & 0.292 & 50.01 & 1 & $-1.118$ \\
GraphSAGE+GRU, ep150 & 0.715 & 0.525 & 32.77 & 12 & $-5.800$ \\
\bottomrule
\end{tabular}
\end{table}

SepGFNO transfers with little loss across geographically and topologically distinct regions (Table~\ref{tab:river}). The Yangtze-trained model reaches median NSE $0.998$ and KGE $0.983$ on the Mississippi operational evaluation; its lower-decile NSE is $0.985$. The trained model's median full-flow and high-flow relative $L_2$ errors are $0.0280$ and $0.0240$, versus $0.5011$ and $0.5045$ for the anchor-only control (Appendix~\ref{sec:river_appendix}). Thus the physical anchors alone do not account for the learned model's magnitude and high-flow accuracy. Both anchored variants have 1 h median peak-timing error, so timing skill cannot be attributed entirely to learning. SepGFNO learns a basin-independent mapping from local reach attributes (slope, length, roughness, hydraulic geometry, upstream area) to routing coefficients; topology enters through the fixed propagation matrix $\mathbf{N}$. The reach-wise maps and drainage-area summaries in Appendix Figure~\ref{fig:river_map} show that strong transfer spans headwaters and mainstem reaches, with a somewhat wider error spread in intermediate tributaries.

\paragraph{Baselines and efficiency.}
The stabilized sparse GraphSAGE+GRU baseline supports basin-wide evaluation but has lower transfer accuracy (Table~\ref{tab:river}). In a controlled one-epoch workload on a Tesla V100S-PCIE-32GB, SepGFNO FFT-IIR takes 72 s and 2.5 GB peak allocated GPU memory, versus 3{,}121 s and 5.4 GB for the same architecture with sequential IIR, and 396 s and 24.2 GB for stabilized GraphSAGE+GRU (Appendix~\ref{sec:sage_stabilization}). The FFT version is $43.3\times$ faster than its sequential control and $5.5\times$ faster than GraphSAGE+GRU for this workload; these are not matched-accuracy convergence comparisons.

For the river setting we use a closed-form Muskingum--Cunge spectral solver (\S\ref{sec:method}) rather than an $L$-layer FNO stack: the recurrence is known and exact under the linearity assumption, so a learned spectral kernel would only discard this physical prior. The $L$-layer design is appropriate for PDE solving, where the operator is not known in closed form. SepTransolver-style learned attention is in principle applicable but practically infeasible here: $T \approx 10^3$ hourly steps and sparse tree adjacency over $5\!\times\!10^4$ reaches would force either many autoregressive windows or prohibitive memory; Physics-Attention's slice tokens (typically 32) cannot meaningfully summarize $5\!\times\!10^4$ reaches spanning headwaters, mainstem, and tributaries. The efficiency advantage of time-parallel solvers also scales proportionally to the number of time steps.
Such computational efficiency helps enable continental- and global-scale streamflow modeling across all river reaches, supporting the expanding scope of physics-embedded hydrologic models~\citep{song2025national,ji2025global,yang2026hourly}.

\Needspace{28\baselineskip}
\subsection{PDE Solving on Irregular Geometries}
\label{sec:pde}
\begin{wraptable}{r}{0.38\textwidth}
\centering
\small
\setlength{\tabcolsep}{6pt}
\caption{Original triangle+notch cross-geometry NSE at 4g, matched-aug protocol (Appendix~\ref{sec:nse_data}); separate from the five-target scaling sweep. \emph{External}: published methods adapted (Appendix~\ref{sec:related_benchmarks}). \emph{Ablation}: variants replacing SepFNO's spatial mixer. SepTransolver no-aug reaches $0.046$.}
\label{tab:pde_baselines}
\begin{tabular}{@{}lcc@{}}
\toprule
Method & Params & $L_2$ \\
\midrule
\textbf{SepFNO (ours)} & \textbf{454K} & \textbf{0.041} \\
\midrule
\multicolumn{3}{@{}l}{\emph{External (adapted)}} \\
SepTransolver & 630K & 0.045 \\
MeshGraphNet & 419K & 0.248 \\
\midrule
\multicolumn{3}{@{}l}{\emph{Ablation (ours)}} \\
GraphSAGE\_FNO & 307K & 0.060 \\
MLP\_FNO & 196K & 0.110 \\
GCN\_FNO & 48K & 0.279 \\
\bottomrule
\end{tabular}
\end{wraptable}
SepFNO, our first Navier--Stokes solver, is trained across fundamentally different, non-diffeomorphic geometries ($\sim$1{,}200--1{,}700 mesh nodes; Appendix~\ref{sec:nse_data}) and evaluated zero-shot on a held-out geometry. While operator learning on irregular geometries is an active area with related benchmarks (PDEBench~\citep{takamoto2022pdebench}, SURF~\citep{kunzli2024surf}, MGN cross-geometry studies~\citep{schmocker2024meshgraphnets}), we are not aware of a standardized benchmark that specifically tests how cross-geometry error scales with the number and diversity of training geometries (Appendix~\ref{sec:related_benchmarks} surveys these). Our second solver adapts Transolver's Physics-Attention~\citep{wu2024transolver} as the spatial mixer within our protocol, yielding \textbf{SepTransolver} (Appendix~\ref{sec:transolver_aug_appendix}). Canonical Transolver targets only single-geometry training; SepTransolver is our construction with five protocol changes (time-windowed channel I/O, graph-free boundary features, multi-mesh shared-weight batching, removal of scalar-time conditioning, stability-tuned LR; Appendix~\ref{sec:transolver_aug_appendix}). We also adapt MeshGraphNets~\citep{pfaff2021meshgraphnets} as an external mesh-based baseline. We additionally construct three ablation baselines---\textbf{MLP\_FNO}, \textbf{GCN\_FNO}, and \textbf{GraphSAGE\_FNO}---by replacing SepFNO's spatial Chebyshev mixer with, respectively, no spatial coupling, a learned GCN aggregator~\citep{kipf2017semi}, and a GraphSAGE 1-hop aggregator~\citep{hamilton2017inductive}, while retaining the same FNO temporal structure. These ablation baselines isolate the contribution of the fixed Chebyshev mixer.

\begin{wrapfigure}{R}{0.46\textwidth}
\centering
\includegraphics[width=\linewidth]{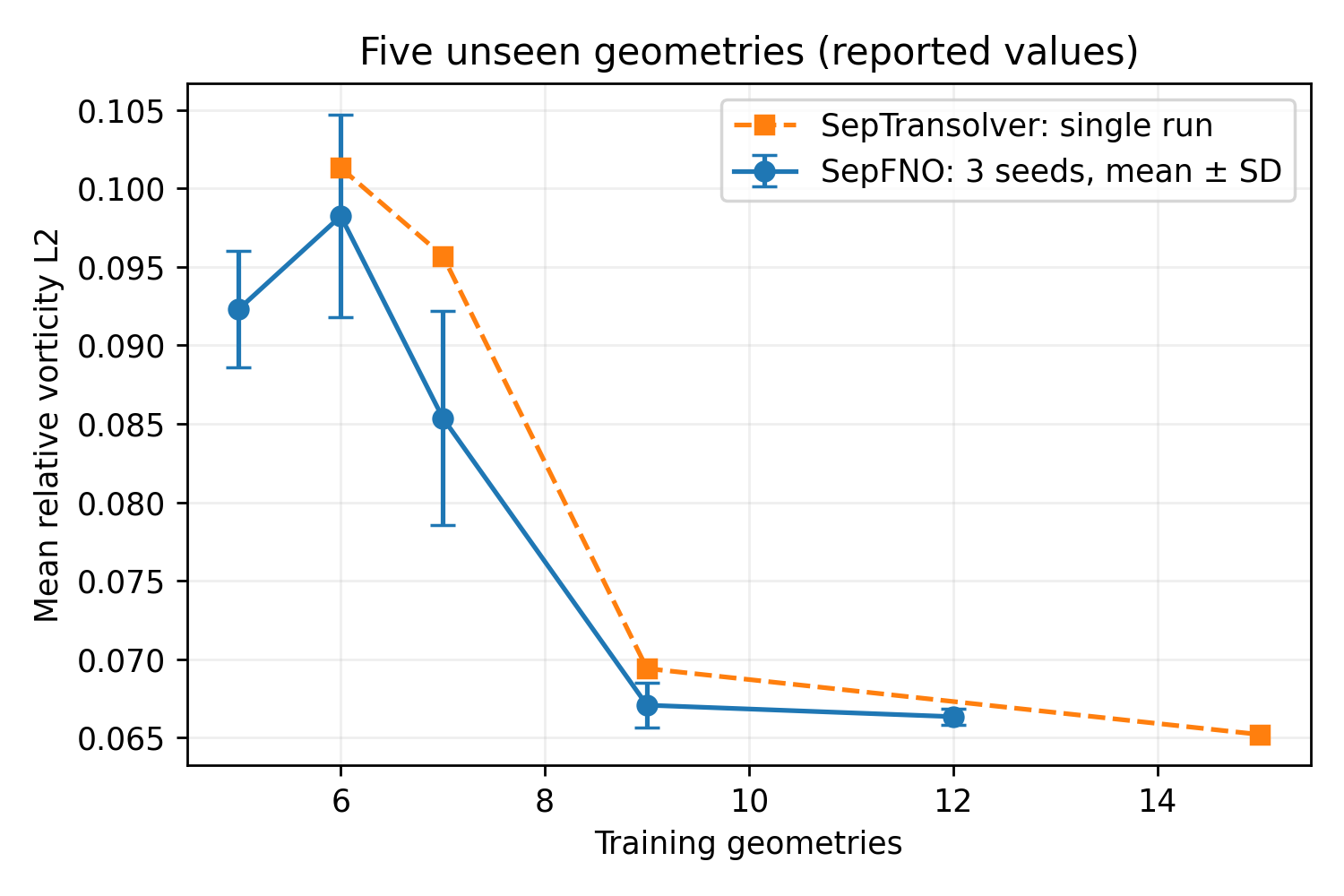}
\caption{Cross-geometry knowledge accumulation across two space-time-factorized families on five unseen geometries. SepFNO (fixed-spectral mixer) shows three-seed mean $\pm$ SD; SepTransolver (learned-attention mixer) is single-run. Pools differ by size and composition.}
\label{fig:scaling}
\end{wrapfigure}

The PDE results have two complementary parts. We first measure how each leak-closing step from \S\ref{sec:method} contributes with 4 training geometries (4g, Table~\ref{tab:pde_ablation}); this isolates per-component effects at a fixed protocol where all baselines train stably. We then examine how cross-geometry error evolves as more training geometries are added (Figure~\ref{fig:scaling}), testing whether the principle supports \emph{knowledge accumulation} across geometries.

\paragraph{Effect of leak control and architectural choices.}
Closing the Level-2 (coordinate) and Level-3 (distribution) leaks of \S\ref{sec:method}, together with the Chebyshev mixer and a longer training schedule, more than halves the cross-geometry error from $L_2{=}0.092$ down to $0.041$ at four training geometries, bringing it within $\sim$2$\times$ of the test-only oracle (Table~\ref{tab:pde_ablation}). Each step contributes measurably and no single intervention dominates: absolute coordinates are the largest single leak (removing $(x,y)$ reduces error by 27\%), the Chebyshev series adds an additional 10\%, and the augmentations combined remove a further 26\% by reducing geometry-specific signatures that persist in boundary features and vorticity magnitudes after coordinate identity is removed. These component-level effects explain the strength of the benchmarks reported below.

\begin{table}[ht]
\centering
\caption{PDE ablation at 4g: each row modifies the previous configuration.}
\label{tab:pde_ablation}
\small
\begin{tabular}{@{}lcc@{}}
\toprule
Configuration & Cross-geo $L_2$ & Main effect \\
\midrule
SepFNO + position $(x,y)$ & 0.092 & coordinate/geometry identity leak \\
$-$ remove position & 0.068 ($-27\%$) & absolute-coordinate memorization \\
+ Chebyshev $T_0 \ldots T_6(\mathbf{A})$ & 0.061 ($-10\%$) & richer fixed spatial mixing \\
+ rotation aug + BF dropout & 0.056 ($-8\%$) & boundary-shape memorization \\
+ $\omega$-scale augmentation & 0.046 ($-18\%$) & magnitude--geometry correlation \\
+ longer schedule (300ep+ft) & \textbf{0.041} ($-11\%$) & convergence/refinement \\
\midrule
Oracle (test-only, ref.) & $\sim$0.018 & --- \\
\bottomrule
\end{tabular}
\end{table}

\begin{figure}[h]
\centering
\includegraphics[width=0.70\textwidth]{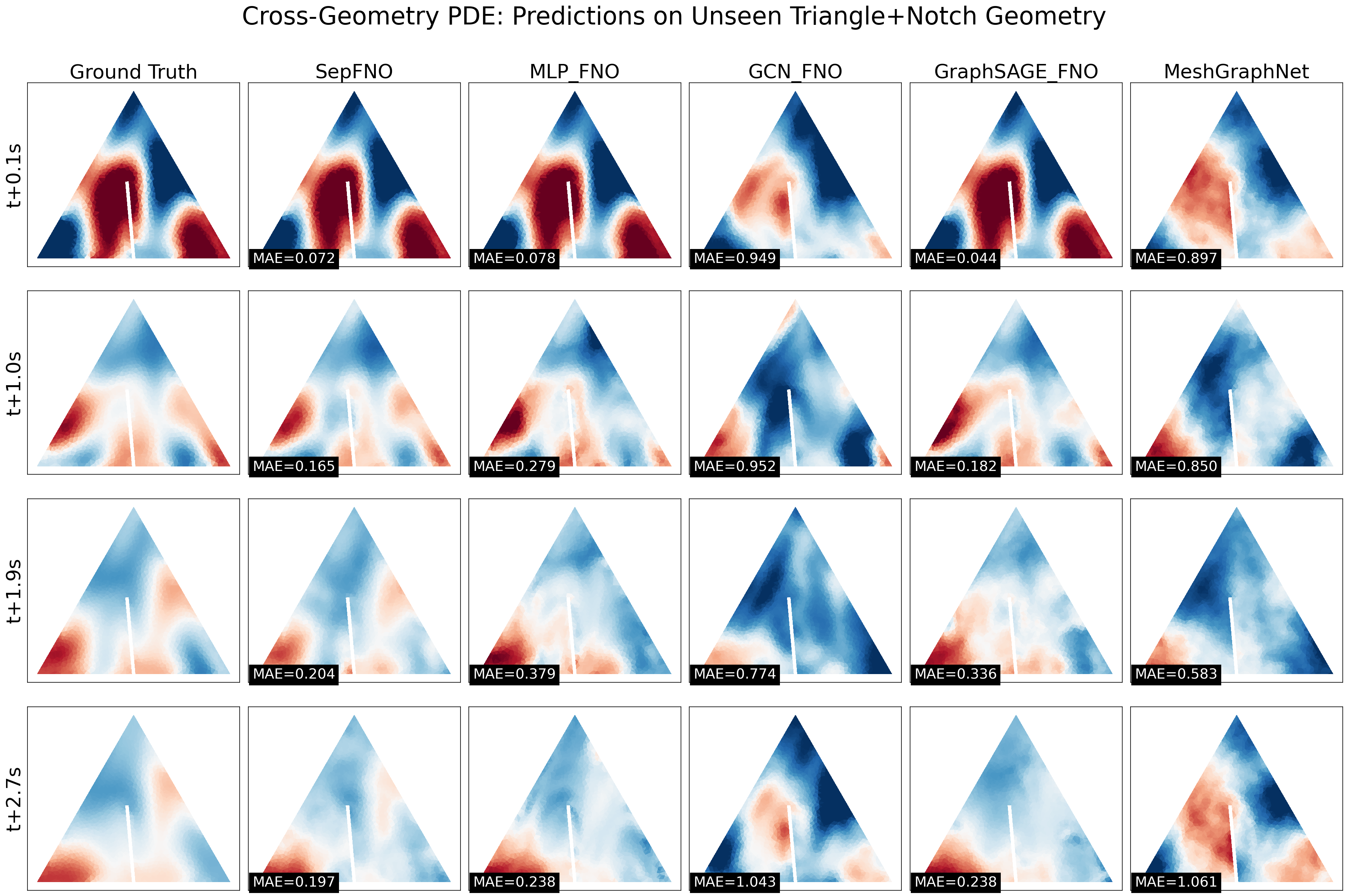}
\caption{Representative 20-step vorticity rollout on an unseen triangle+notch geometry. SepFNO preserves the main vorticity structures, while weaker baselines produce visibly distorted fields.}
\label{fig:pde_predictions}
\end{figure}

Figure~\ref{fig:pde_predictions} illustrates the quantitative comparison with a 20-step vorticity rollout on the held-out triangle-with-notch mesh: SepFNO retains the principal flow structures while weaker baselines visibly distort them.

\paragraph{Cross-architecture comparison (Table~\ref{tab:pde_baselines}).}
Both SepFNO ($0.041$) and SepTransolver ($0.045$) outperform all structural-ablation baselines by $1.5$--$7\times$, and their agreement to within $\sim$10\% provides empirical evidence that the factorization principle accommodates more than one spatial-mixer family (fixed-spectral and learned-attention here). The remaining baselines fail in ways predicted by the leakage framework above: \textbf{GCN\_FNO} ($0.279$) uses learned graph-coupled aggregation, a textbook Level-1 leak; \textbf{MeshGraphNet} ($0.248$) uses relative-position edge features that are standard in single-mesh settings but encode geometry identity under cross-geometry transfer (a Level-2 leak embedded in the spatial mixer itself); \textbf{GraphSAGE\_FNO} ($0.060$) uses 1-hop mean aggregation that satisfies the leakage constraints but is expressively weaker than the Chebyshev $K{=}6$ mixer; \textbf{MLP\_FNO} ($0.110$) has no spatial coupling at all and therefore cannot propagate information across nodes. The MLP\_FNO vs.\ GCN\_FNO/MeshGraphNet comparison surfaces a non-obvious lesson: \emph{mis-specified learned spatial coupling is worse for cross-geometry transfer than no spatial coupling at all.}

\paragraph{Knowledge accumulation: quantity and relatedness.}
Figure~\ref{fig:scaling} tests \emph{cross-geometry knowledge accumulation across two space-time-factorized model families}: fixed-spectral SepFNO and learned-attention SepTransolver. On the same five unseen targets, the final three-seed SepFNO sweep reduces mean relative vorticity error from $0.0923\pm0.0037$ at 5g to $0.0663\pm0.0005$ at 12g (about 28\%); a separate single-run SepTransolver sweep falls from $0.1013$ at 6g to $0.0652$ at 15g (Appendix Table~\ref{tab:scaling_new}). The common direction across different spatial mixers supports accumulation beyond one architecture or one test geometry. It is an empirical trend across sampled pools, not a monotone law: SepFNO worsens at 6g and improves little from 9g to 12g. Equal-size SepTransolver controls also show that pool composition matters: adding a sinusoidal channel improves mean error from $0.0994$ to $0.0754$, whereas adding an hourglass worsens it to $0.1102$. Matched target-only oracle gaps still vary by geometry (Appendix~\ref{sec:nse_data}).

The six-target check in Appendix Table~\ref{tab:pde_six} uses fixed 4g checkpoints without target adaptation. SepFNO has the lowest vorticity and reconstructed-velocity $L_2$ on every held-out geometry among the five benchmarked models; its across-target means are $0.0482/0.0283$, versus $0.0674/0.0398$ for the next-best GraphSAGE\_FNO. Pressure is absent from this vorticity formulation. Per-node analysis finds larger boundary errors (Appendix~\ref{sec:boundary_error}). As a public-data test beyond our generated geometries, we also train SepFNO on the unsteady Schmöcker cylinder-flow benchmark~\citep{schmocker2024meshgraphnets} and transfer zero-shot to the unseen stretched cylinder. In the matched laminar subset ($Re<47$, $n=11$), its median 20-step velocity RMSE is $76.0\times10^{-3}$, versus $163.0\times10^{-3}$ for native MeshGraphNet (Appendix Table~\ref{tab:schmocker}). This supports transfer in a separate geometry family, although the MeshGraphNet checkpoint we evaluated is undertrained; the broader Reynolds-number results and this fairness limit are given in the appendix.

\subsection{Traffic Speed Forecasting}
\label{sec:traffic}

Traffic forecasting tests a learned-dynamics setting where no physical recurrence is available. We use the DCRNN benchmark protocol~\citep{li2018dcrnn}: models are trained on METR-LA (207 sensors) and deployed zero-shot on PEMS-BAY (325 sensors). Unlike the river and PDE cases, the protocol uses a short prediction horizon (12 steps in, 12 steps out), where FFT-in-time offers little wall-clock advantage over a recurrent core; temporal dynamics are therefore learned by a shared GRU, while the factorization principle is enforced at the spatial side. Spatial coupling is introduced through fixed multi-hop diffusion over the target road graph, with node-wise hop weights predicted from local sensor statistics rather than from sensor identities. The choice of temporal solver (FFT versus GRU) is domain-driven: the principle constrains where graph topology may enter, not which temporal solver the local dynamics use. At deployment, SepGRU-v2 computes local hop weights from unlabeled target-city sensor statistics; no target-city labels are used for training or adaptation.

\begin{table}[ht]
\centering
\caption{Traffic METR-LA$\to$PEMS-BAY zero-shot transfer (DCRNN 12-step protocol; 30-min MAE on PEMS-BAY). \emph{NodeEmbedGRU}/\emph{EntangledGRU} are our re-implementations of the AGCRN~\citep{bai2020agcrn}/DCRNN~\citep{li2018dcrnn} cross-city paradigms. SepGRU-v2 robustness: $2.16 \pm 0.01$ over 3 seeds.}
\label{tab:traffic}
\small
\setlength{\tabcolsep}{4.5pt}
\renewcommand{\arraystretch}{1.08}
\begin{tabular}{@{}lccc@{}}
\toprule
Model & Transfer MAE & Oracle gap & Transfer factor tested \\
\midrule
\textbf{SepGRU-v2 + dropout (ours)} & \textbf{2.13} & \textbf{+12.9\%} & feature-conditioned diffusion + regularization \\
SepGRU-v2 (ours) & 2.16 & +14.3\% & feature-conditioned diffusion weights \\
EntangledGRU (DCRNN-style) & 2.18 & +16.8\% & shared graph-in-recurrence, no node IDs \\
SepGRU-v1 (ours, no learned $\alpha$) & 2.23 & +19.3\% & fixed hop weights, no local adaptation \\
NodeEmbedGRU (AGCRN-style) & 2.91 & +55.9\% & source-node embeddings \\
\bottomrule
\end{tabular}
\end{table}

The traffic results isolate a Level-1 leak. NodeEmbedGRU follows standard practice~\citep{bai2020agcrn} in assigning each sensor a learnable embedding vector indexed by sensor ID; these embeddings are graph-specific by construction, so the embedding for ``LA sensor 47'' is meaningless on PEMS-BAY and the oracle gap rises to $55.9\%$. Replacing them with diffusion weights predicted from \emph{local sensor statistics} (mean speed, variance, congestion fraction) restores transferability: the gap drops to $14.3\%$ for SepGRU-v2 and $12.9\%$ with dropout in encoder and coefficient head. SepGRU-v1, with fixed diffusion and no per-node adaptation, is worse ($+19.3\%$), so some local adaptation helps---as long as it is predicted from transferable features rather than learned per identity. EntangledGRU, which places graph diffusion inside the recurrent update without node-specific weights, remains close to SepGRU-v2 ($2.18$ vs. $2.16$): cross-city transfer breaks most clearly when learned parameters are tied to source sensors, not from graph use inside the recurrence per se.

\subsection{Breadth Check and Negative Controls}
\label{sec:negative}

\textbf{Citation networks} (breadth check beyond spatiotemporal): on GDABench~\citep{gdabench2024} (DBLP, ACM, Citation; Appendix~\ref{sec:citation_appendix}), a SIGN-style pre-smoothed model with graph-free MLP heads reaches 64.8\% average Macro-F1 vs.\ 63.7\% for a 3-layer GCN across six source$\to$target pairs---a modest gain confirming the rule applies beyond temporal dynamics. As negative controls, \textbf{Power flow} (IEEE 24/39/118-bus, cross-system transfer): the admittance matrix encodes line impedances that enter Kirchhoff's laws directly---the graph \emph{is} the physics. Graph models incur $+500\%$ transfer gaps; a graph-free MLP on local injection features increases error by $12\%$. \textbf{Airport routes} (GDABench Brazil/Europe/USA): node features are one-hot degree encodings (graph-derived, not physical), so SepGNN (44.3\% F1) falls below MLP (54.5\%) and GCN (50.2\%) (applicability checklist in Appendix~\ref{sec:applicability_table}).

\section{Conclusion}
\label{sec:conclusion}

We have argued that separating temporal evolution from graph propagation changes what kinds of generalization and scaling become possible for graph-based dynamical systems. Once topology no longer participates in the recurrent update, time-parallel solvers become applicable on irregular graphs---a regime classical time iteration struggles with. The contribution is the \textbf{factorization protocol} (graph-free features, time-windowed I/O, fixed or graph-free spatial mixing, architecture-tuned leak control), not a new primitive layer. Empirically (Fig.~\ref{fig:scaling}), the same protocol unlocks both fixed-spectral (SepFNO/Chebyshev) and learned-attention (SepTransolver/Physics-Attention) mixers, with cross-geometry error decreasing in both training-geometry quantity and test-relatedness. The three domains exemplify distinct spatial-mixer choices appropriate to the regime---MC/FFT-IIR with fixed upstream propagation for rivers, Chebyshev or Physics-Attention mixing for PDEs, and GRU dynamics with fixed multi-hop diffusion for traffic.

Several limitations remain. The PDE cross-geometry gap is reduced by data scaling but not yet closed, and a linear temporal kernel cannot capture state-dependent spatial coupling without additional regimes or layer stacking; physics-informed extensions---such as embedding a differentiable Poisson solver or imposing soft sensitivity constraints around boundaries---are a natural next step.

Two directions stand out: scaling laws for graph dynamical systems---how cross-graph error decreases with the number, diversity, and similarity of training graphs---and opening a path toward foundation models for physical simulation, where many graph instances are pooled and a single model generalizes across them. We hope this paper provides one structural argument for why such an objective is reachable, and a concrete starting point for it.

\section*{Funding}
This work is supported by the Cooperative Institute for Research to Operations in Hydrology (CIROH) through the National Oceanic and Atmospheric Administration (NOAA) Cooperative Agreement (Grant NA22NWS4320003) and National Aeronautics and Space Administration (NASA) grant 80NSSC24K1646.

\section*{Competing Interests}
C.S. has financial interests in HydroSapient, Inc., a company which could potentially benefit from the results of this research. These interests have been reviewed by The Pennsylvania State University in accordance with its individual conflict of interest policy, for the purpose of maintaining the objectivity and the integrity of research at The Pennsylvania State University.


\bibliographystyle{plainnat}
\bibliography{references}

\appendix

\renewcommand{\thefigure}{S\arabic{figure}}
\renewcommand{\thetable}{S\arabic{table}}
\setcounter{figure}{0}
\setcounter{table}{0}

\section{Instantiations of the Factorization Principle}
\label{sec:architectures_appendix}

Table~\ref{tab:architectures} summarizes how the factorization template is instantiated across the five models in the paper. The spatial mixer is the modular slot whose choice depends on the regime (Long horizons and large sparse graphs favor fixed propagation for scalability; smaller irregular meshes can afford learned graph-free attention).

\begin{table}[!htbp]
\centering
\caption{Instantiations across domains. The spatial mixer is the modular slot; choice depends on regime (long $T$ + sparse adjacency favors fixed-spectral; dense local mesh admits attention).}
\label{tab:architectures}
\footnotesize
\setlength{\tabcolsep}{3pt}
\begin{tabular}{@{}p{2.4cm}p{3.3cm}p{3.3cm}p{2.0cm}r@{}}
\toprule
Domain (model) & Local features $\to$ coefficients & Spatial mixer (slot) & Temporal solver & Params \\
\midrule
River (SepGFNO) & Reach attrs $\to$ routing coeffs & Upstream sum $\mathbf{N}\mathbf{Q}$ (fixed) & FFT-IIR & 27K \\
PDE (SepFNO) & Boundary feats $\to$ mixing weights & Chebyshev $T_k(\mathbf{A})\mathbf{h}$ (fixed) & FFT & 454K \\
PDE (SepTransolver) & Boundary feats $\to$ slice tokens & Physics-Attention (learned, graph-free) & windowed I/O & 630K \\
Traffic (SepGRU) & Sensor stats $\to$ hop weights & $\sum\alpha_k \mathbf{A}^k\mathbf{h}$ (fixed) & GRU & 51K \\
Citation (SepHybrid) & Text feats $\to$ hidden state & Pre-smoothing $\mathbf{A}^K\mathbf{x}$ (fixed) & -- & -- \\
\bottomrule
\end{tabular}
\end{table}

\section{Applicability Checklist}
\label{sec:applicability_table}

Table~\ref{tab:applicability} summarizes which of the six domains we evaluate satisfy the two necessary conditions of \S\ref{sec:applicability} (graph-independent features; graph is connectivity only) and whether the separation principle works empirically. The four positive domains (river, PDE, traffic, citation) satisfy both criteria; the two negative controls (power flow, airport networks) violate one each.

\begin{table}[!htbp]
\centering
\caption{Applicability checklist across the six domains evaluated.}
\label{tab:applicability}
\small
\begin{tabular}{@{}lcccccc@{}}
\toprule
Criterion & River & PDE & Traffic & Citation & Power & Airport \\
\midrule
Graph-independent features? & Yes & Yes & Yes & Yes & Partial & No \\
Graph is connectivity only? & Yes & Yes & Yes & Yes & No & Yes \\
Separation works? & Yes & Yes & Yes & Yes & No & No \\
\bottomrule
\end{tabular}
\end{table}

\section{Additional River Results}
\label{sec:river_appendix}

\begin{figure}[!htbp]
\centering
\includegraphics[width=0.88\textwidth]{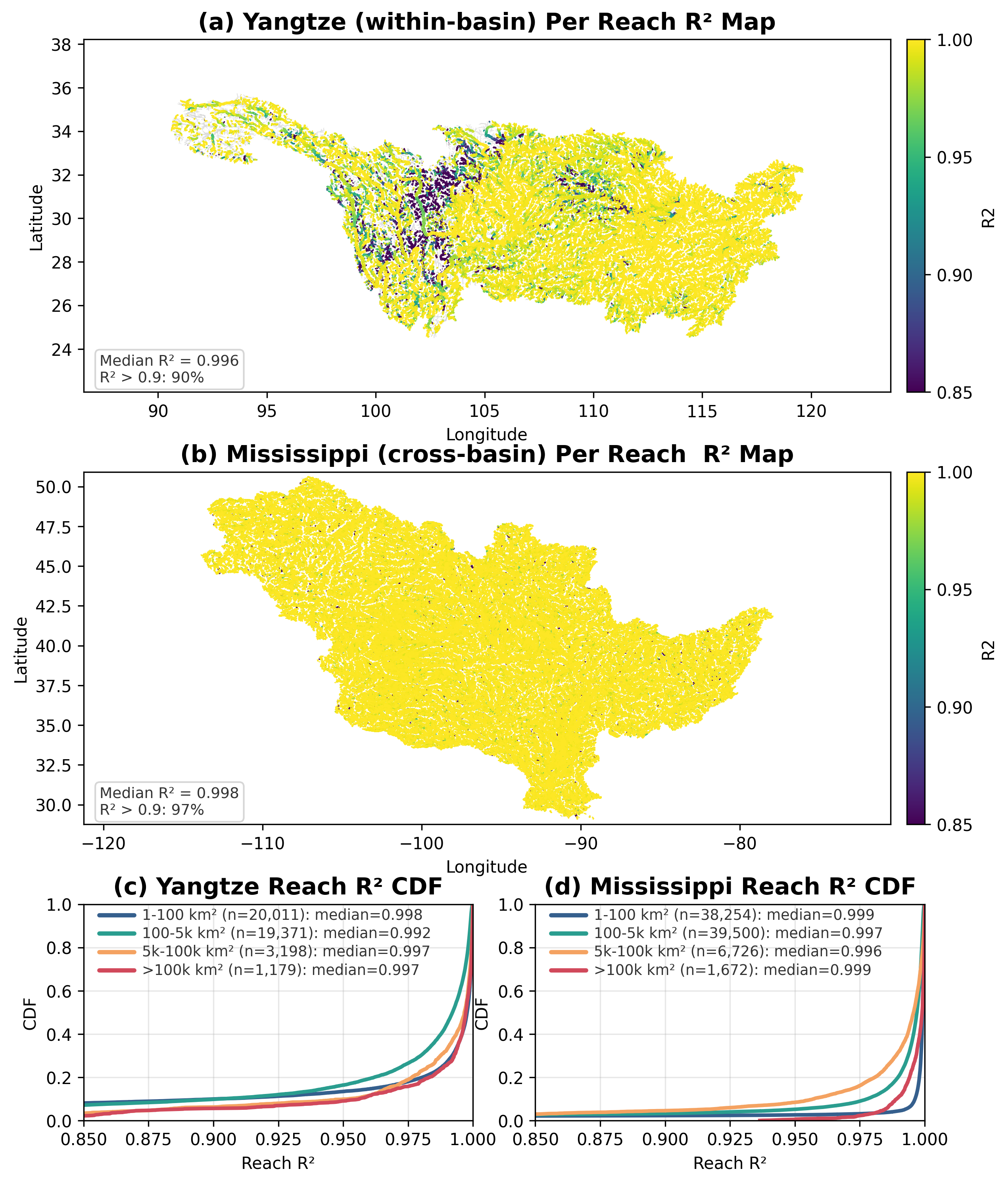}
\caption{Reach-wise performance using Yangtze-trained SepGFNO weights. Per-reach $R^2$ maps for (a) Yangtze and (b) Mississippi, and $R^2$ by upstream drainage area for (c) Yangtze and (d) Mississippi. High accuracy spans small and large drainage areas, with somewhat wider spread in intermediate tributaries.}
\label{fig:river_map}
\end{figure}

Figure~\ref{fig:river_map} shows that high reach-wise $R^2$ extends across both basins and from headwaters to mainstem reaches. The intermediate drainage-area bins have a somewhat broader error distribution, so the strong basin-wide median does not imply uniform accuracy at every reach.

The common-cohort metrics in Table~\ref{tab:river} intersect the outputs of all three models at 86{,}133 active reaches. After $t_0=200$, each reach has 2{,}400 hourly steps; only finite NSE values and target coefficient of variation above 0.05 are retained. Peak error uses the target window maximum and searches the predicted peak within $\pm24$ h. The anchor-only control bypasses learned coefficient corrections but retains the five-regime architecture with untrained components; it is not a pure deterministic Muskingum--Cunge solver. Its 1 h timing median matches the trained model, while magnitude and shape differ substantially. Operational trunk injection nodes receive reference flows in both SepGFNO and GraphSAGE+GRU evaluations; these results therefore measure boundary-conditioned transfer, not fully closed-loop upstream propagation.

\begin{table}[!htbp]
\centering
\caption{Mississippi common-cohort relative $L_2$: median [P10, P90] across 86{,}133 reaches. High flow restricts each reach to times at or above its target 98th percentile.}
\label{tab:river_l2}
\small
\begin{tabular}{@{}lcc@{}}
\toprule
Model & All times & High flow \\
\midrule
SepGFNO, trained & 0.0280 [0.0120, 0.0809] & 0.0240 [0.0088, 0.0686] \\
SepGFNO, anchor-only & 0.5011 [0.4998, 0.5194] & 0.5045 [0.5000, 0.5477] \\
GraphSAGE+GRU, ep150 & 0.3606 [0.0700, 1.5143] & 0.3410 [0.0415, 0.8570] \\
\bottomrule
\end{tabular}
\end{table}

\section{Boundary Error Localization (PDE)}
\label{sec:boundary_error}

\begin{figure}[h]
\centering
\includegraphics[width=0.95\textwidth]{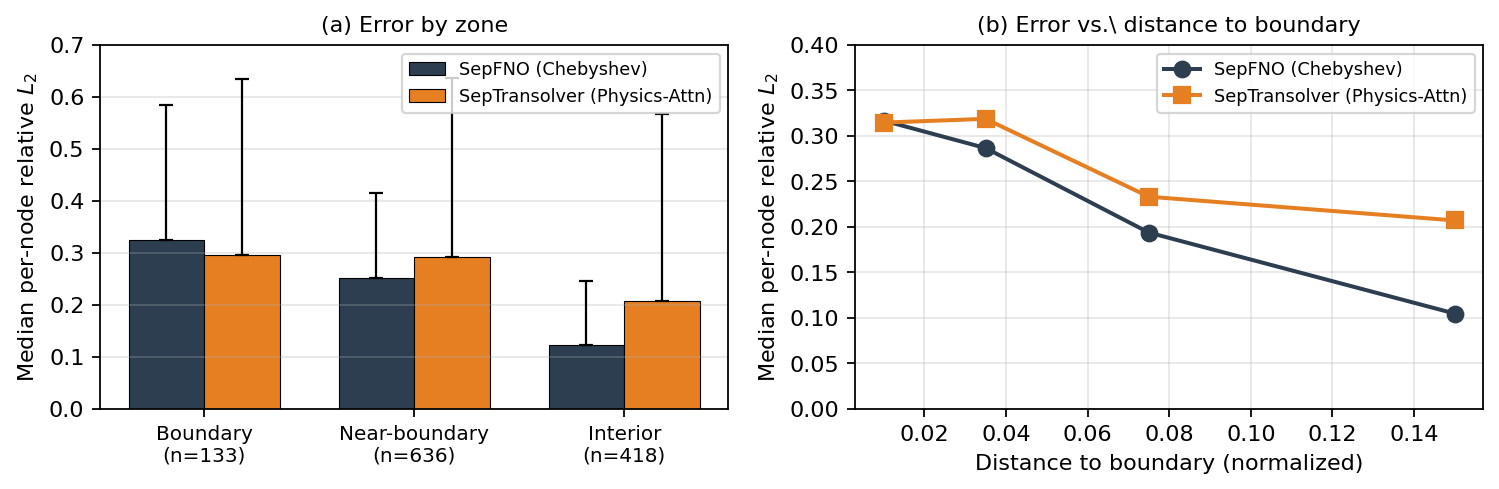}
\caption{Cross-geometry error patterns differ structurally between the two spatial mixers. (a)~Median per-node relative $L_2$ by zone (whisker = 75th percentile). SepFNO's Chebyshev polynomial mixing concentrates error at the boundary (boundary/interior ratio $\sim$2.7$\times$); SepTransolver's Physics-Attention slice tokens distribute error far more uniformly (ratio $\sim$1.4$\times$). (b)~Median error vs.\ normalized distance to boundary: SepFNO drops $\sim$3$\times$ from boundary to deep interior (0.32$\to$0.10), while SepTransolver drops only $\sim$1.5$\times$ (0.31$\to$0.21). Both architectures struggle most near boundaries, but the magnitude of locality bias differs: fixed local-polynomial mixing is locality-sensitive in its errors; global-attention mixing spreads error uniformly across the mesh.}
\label{fig:boundary_error}
\end{figure}

Figure~\ref{fig:boundary_error} shows where cross-geometry residual error concentrates. We partition the test mesh (triangle+notch, 1187 nodes) into three zones: \emph{boundary} (nodes on the mesh boundary, $n=133$), \emph{near-boundary} (within $\sim$2 mesh layers of the boundary, $n=636$), and \emph{interior} (the rest, $n=418$). Per-node relative $L_2$ is averaged over 30 evaluation samples on the 4-geometry checkpoints of both SepFNO (matched-aug) and SepTransolver (no aug, the per-architecture best).

\section{Architecture Details}
\label{sec:arch_details}

\subsection{SepGFNO: River Routing}
\label{sec:sepgfno_detail}

\paragraph{Physical basis: Muskingum--Cunge routing.}
The Muskingum--Cunge (MC) method~\citep{bindas2024dmc} is a finite-difference solution to the one-dimensional continuity equation for open-channel flow, routing discharge through a river network reach by reach. For a single reach, the MC equation relates outflow $Q$ at time $t$ to inflow $I$ from upstream and lateral inflow $q'$ from the local catchment:
\begin{equation}
    Q_i(t) = c_{1,i}\, I_i(t) + c_{2,i}\, I_i(t{-}1) + c_{3,i}\, Q_i(t{-}1) + c_{4,i}\, q'_i(t)
    \label{eq:mc}
\end{equation}
where the coefficients $c_1$--$c_4$ encode the reach's hydraulic properties. These coefficients are derived from Manning's equation and channel geometry. Following the differentiable MC framework ($\delta$MC)~\citep{bindas2024dmc,ji2025global}, channel flow velocity is computed from Manning's equation:
\begin{equation}
    v = \frac{1}{n}\, d^{2/3}\, S_0^{1/2}, \quad
    d = \left(\frac{Q\, n\, (q_s{+}1)}{p_s\, S_0^{1/2}}\right)^{\!\frac{3}{5+3q_s}}
    \label{eq:manning}
\end{equation}
where $n$ is Manning's roughness, $S_0$ is bed slope, $d$ is flow depth, and $p_s, q_s$ are hydraulic geometry parameters relating top width $w = p_s\, d^{q_s}$ to depth~\citep{bindas2024dmc}. Wave celerity $c = \tfrac{5}{3} v$ (clamped to $[0.3, 15]$~m/s) determines the wave travel time $K_w = \Delta x / c$, from which the MC coefficients follow:
\begin{equation}
    c_1 = \frac{\Delta t - 2K_w X}{D}, \quad
    c_2 = \frac{\Delta t + 2K_w X}{D}, \quad
    c_3 = \frac{2K_w(1{-}X) - \Delta t}{D}, \quad
    c_4 = \frac{\Delta t}{D}
    \label{eq:mc_coeffs}
\end{equation}
where $D = 2K_w(1{-}X) + \Delta t$, $X = 0.3$ is the storage coefficient, and $\Delta t = 3600$~s. Mass conservation requires $c_1 + c_2 + c_3 + c_4 = 1$.

In $\delta$MC, the coefficients are computed sequentially---each reach routes flow from its immediate upstream neighbor, proceeding from headwaters to outlet in topological order. This sequential formulation, while physically accurate, cannot exploit GPU parallelism and requires $O(D \cdot T)$ sequential operations for a network of diameter $D$.

\paragraph{Matrix reformulation for GPU parallelism.}
SepGFNO reformulates the reach-by-reach routing as simultaneous matrix operations over the entire network. The upstream connectivity is encoded in a binary matrix $\mathbf{N} \in \{0,1\}^{m \times m}$ where $N_{ij} = 1$ if reach $j$ is an immediate upstream neighbor of reach $i$ (with appropriate normalization for confluence nodes). The inflow to each reach is then:
\begin{equation}
    \mathbf{I}(t) = \mathbf{N}\, \mathbf{Q}(t)
    \label{eq:matrix_routing}
\end{equation}
where $\mathbf{Q}(t) \in \mathbb{R}^m$ is the vector of all reach discharges at time $t$. Since $\mathbf{N}$ is strictly lower-triangular when nodes are sorted by upstream area (headwaters first), the matrix power $\mathbf{N}^D = \mathbf{0}$ for $D$ equal to the graph diameter. Starting from an initial guess (e.g., headwater-only routing), iterating Eq.~\eqref{eq:mc} with $I_i(t) = (\mathbf{N}\mathbf{Q}(t))_i$ converges to the exact sequential solution in $D$ iterations. Each iteration is a single sparse matrix--vector multiply, fully parallelizable on GPU.

\paragraph{Step 1 (Encoder).} An MLP maps local reach attributes $\mathbf{a}_i = (S_0, \Delta x, n, q_s, p_s, \text{uparea}, \ldots) \in \mathbb{R}^{N_\text{feat}}$ to a hidden vector $\mathbf{h}_i = \text{MLP}_\text{enc}(\mathbf{a}_i) \in \mathbb{R}^d$, shared across all reaches and basins.

\paragraph{Step 2 (Coefficient prediction with physics anchors).}
For each of $K$ flow regimes (corresponding to quantiles of the flow distribution: low, medium, high), physics-anchored MC coefficients $\tilde{c}_{1,k}, \ldots, \tilde{c}_{4,k}$ are computed from Manning's equation (Eq.~\eqref{eq:manning}--\eqref{eq:mc_coeffs}) using representative discharge values. These anchors are then modulated by learned correction heads:
\begin{equation}
    c_{2,k,i} = \sigma(\mathbf{W}_2 \mathbf{h}_i)_k \cdot 2 \cdot \tilde{c}_{2,k,i}, \quad
    c_{1,k,i} = 1 - c_{2,k,i} - c_{3,k,i} \quad \text{(mass conservation)}
\end{equation}
When the MLP output is zero, $\sigma(0) \cdot 2 = 1$ and the coefficients match the physics exactly. The MLP learns corrections to account for sub-grid heterogeneity and model structural error. All MLP weights are graph-independent---the same parameterization scheme applies to any river basin.

\paragraph{Step 3 (FFT-IIR temporal processing).}
The MC equation (Eq.~\eqref{eq:mc}) is a first-order IIR (infinite impulse response) filter with feedback coefficient $c_3$. Its impulse response is $h[t] = c_3^t$ for $t \geq 0$. Given an input signal $u_i(t) = c_1 I_i(t) + c_2 I_i(t{-}1) + c_4 q'_i(t)$, the output is a convolution:
\begin{equation}
    Q_{k,i}(t) = (u_i * h_k)[t] = \sum_{\tau=0}^{t} u_i(\tau)\, c_{3,k,i}^{t-\tau}
\end{equation}
This convolution is computed in the frequency domain:
\begin{equation}
    H_k = \text{FFT}(c_{3}^0, c_{3}^1, \ldots, c_{3}^{T-1}), \quad
    \mathbf{Q}_k = \text{IFFT}(\text{FFT}(\mathbf{u}_k) \odot H_k)
    \label{eq:fftiir}
\end{equation}
processing all $T$ timesteps in $O(T \log T)$ with \textbf{no backpropagation through time}. This is the key efficiency gain: standard autoregressive GRU/LSTM baselines require $O(T)$ sequential steps with full BPTT memory.

When an initial condition $Q_i(0) = Q_\text{init}$ is available (warm start), the homogeneous solution $Q_\text{homo}(t) = c_3^{t+1} \cdot Q_\text{init}$ is added to the convolution output analytically.

\paragraph{Step 4 (Iterative spatial propagation).}
Starting from headwater-only solutions (Eq.~\eqref{eq:fftiir} with $I = 0$), we iterate:
\begin{align}
    \mathbf{I}^{(\ell)}(t) &= \mathbf{N}\, \mathbf{Q}^{(\ell-1)}(t) \\
    \mathbf{Q}^{(\ell)} &= \text{FFT-IIR}\big(c_1 \mathbf{I}^{(\ell)} + c_2 \mathbf{I}^{(\ell)}_{t-1} + c_4 \mathbf{q}',\; c_3\big) + \mathbf{Q}_\text{homo}
\end{align}
for $\ell = 1, \ldots, D$ (graph diameter). All $K$ regimes are batched: $\mathbf{Q}^{(\ell)} \in \mathbb{R}^{K \times m \times T}$ is reshaped to $\mathbb{R}^{m \times KT}$ for a single sparse matrix multiply per iteration, then reshaped back. Gradient checkpointing is used during training to bound memory to $O(m \cdot T)$ regardless of $D$.

\paragraph{Step 5 (Regime blending).}
A dilated temporal CNN processes the mean routed discharge (detached from the computation graph) to predict time-varying regime weights:
\begin{equation}
    w_k(t) = \text{softmax}_k\!\left(\text{CNN}\!\left(\frac{\bar{Q}(t)}{\overline{\bar{Q}}}\right)\right), \quad
    Q_\text{final}(t) = \sum_{k=1}^{K} w_k(t)\, Q_k(t)
\end{equation}
The CNN uses dilations $[1, 2, 4, 8]$ with kernel size 7, capturing temporal context at multiple scales. Optional temporal smoothing (average pooling with kernel size 25) reduces high-frequency oscillations in regime weights. The $K$-regime system allows the model to represent nonlinear stage-dependent routing behavior---low flows route slowly (high $c_3$, strong attenuation), while high flows route fast (low $c_3$, near-kinematic wave)---without explicit flow-dependent coefficient update, which would break the FFT-IIR approach.

\paragraph{Subgraph training, leaf transfer, and cascade inference.}
Large river networks can have graph diameters far beyond the number of spatial propagation iterations that is practical during training. We therefore distinguish three evaluation modes. A \emph{leaf subgraph} is a complete upstream drainage subtree terminating at a selected outlet reach; it has no missing upstream boundary and can be routed independently. Leaf-transfer tests train on leaf subgraphs from one basin and evaluate on held-out leaf subgraphs from another basin, measuring whether the learned local routing coefficients generalize across hydroclimatic and geomorphic settings. For full-basin inference, the network is decomposed into bounded-diameter leaves and trunk segments. A \emph{trunk segment} is a mainstem piece whose upstream boundary receives inflow from upstream leaves or earlier trunk segments. In the reported \emph{operational cascade}, segments are processed from upstream to downstream, with reference discharge supplied at trunk injection nodes. This avoids running spatial propagation over the full basin diameter while providing known boundary conditions; the reported metrics do not measure fully closed-loop propagation of upstream prediction error. All experiments use the same learned SepGFNO weights; only the fixed connectivity matrix $\mathbf{N}$, local reach attributes, and boundary inflows change across basins.

\paragraph{Summary.}
The model has 27{,}924 parameters ($K{=}5$ regimes, hidden dim 64). All learned weights (encoder MLP, coefficient heads, temporal CNN) are graph-independent. The only graph-dependent quantity is $\mathbf{N}$, which is a binary connectivity matrix carrying no learned parameters. To deploy on a new basin, one simply constructs $\mathbf{N}$ from the new river network topology and provides the local reach attributes---no retraining, fine-tuning, or adaptation required.

\subsection{SepFNO: PDE on Irregular Meshes (Navier--Stokes)}
\label{app:sepfno}

\textbf{Step 1 (Lift).} Per-node boundary features $\mathbf{b}_i \in \mathbb{R}^6$ (is\_boundary, distance, normal, curvature, density---\emph{no absolute coordinates}) and input vorticity $\omega_i(t)$ are lifted: $\mathbf{h}_i(t) = \mathbf{W}_\text{lift} [\mathbf{b}_i; \omega_i(t)]$.

\textbf{Step 2 (FNO layers, $\times L$).} Each layer applies three operations:

\emph{(a) Temporal FFT:} $\hat{\mathbf{h}}_i(f) = \text{FFT}_t(\mathbf{h}_i(t))$ for frequency modes $f = 0, \ldots, \lfloor T/2 \rfloor$.

\emph{(b) Spectral kernel (graph-free):} $\hat{\mathbf{h}}'(f) = \mathbf{W}_\text{spec}^{(\ell)} [\text{Re}(\hat{\mathbf{h}}(f)); \text{Im}(\hat{\mathbf{h}}(f))]$, where $\mathbf{W}_\text{spec}$ is shared across all nodes and graphs.

\emph{(c) Chebyshev spatial mixing (fixed graph operator):}
\begin{equation}
    T_0(\mathbf{A}) = \mathbf{I}, \;\; T_1(\mathbf{A}) = \mathbf{A}, \;\; T_k(\mathbf{A}) = 2\mathbf{A} T_{k-1}(\mathbf{A}) - T_{k-2}(\mathbf{A})
\end{equation}
\begin{equation}
    \mathbf{s}_i(f) = \mathbf{W}_\text{spat} [T_0 \hat{\mathbf{h}}'_i(f); T_1 \hat{\mathbf{h}}'_i(f); \ldots; T_K \hat{\mathbf{h}}'_i(f)]
\end{equation}
where $\mathbf{W}_\text{spat} \in \mathbb{R}^{d \times (K+1)d}$ is graph-independent. The Chebyshev polynomials use the normalized adjacency (spectrum in $[-1,1]$), eliminating per-graph $\lambda_\text{max}$ rescaling.

\emph{(d) Inverse FFT + residual:} $\mathbf{h}^{(\ell+1)}_i(t) = \text{GELU}(\text{IFFT}_f(\hat{\mathbf{h}}'_i + \mathbf{s}_i) + \mathbf{W}_\text{res}^{(\ell)} \mathbf{h}^{(\ell)}_i(t))$.

\textbf{Step 3 (Projection).} $\hat{\omega}_i(t) = \text{MLP}_\text{proj}(\mathbf{h}_i^{(L)}(t))$.

\paragraph{Training schedule and regularization.}
We train SepFNO with AdamW (initial LR $10^{-3}$, weight decay $10^{-4}$) for 300 epochs under a cosine LR schedule that decays to zero, followed by a fine-tuning phase of up to 100 additional epochs at a constant LR of $10^{-5}$. The fine-tuning phase consistently improves cross-geometry $L_2$ by an additional $\sim$5\% relative to the cosine endpoint, because by the end of the cosine schedule the LR is too small to make further progress while the loss is still slowly decreasing; switching to a constant low LR releases this residual signal without re-entering the noisier high-LR regime. Four regularization mechanisms are used jointly during training, each addressing a distinct leak channel identified in Section~\ref{sec:method}: (i)~\emph{rotation augmentation} applies random $\theta \in [0, 2\pi)$ rotations and reflections to node positions, surface normals, and the vorticity sign, breaking absolute orientation cues; (ii)~\emph{boundary-feature dropout} randomly zeros entire boundary feature columns with probability $p=0.3$, preventing the model from memorizing specific boundary-shape signatures; (iii)~\emph{$\omega$-scale augmentation} multiplies input vorticity by a per-sample lognormal factor $\exp(\pm 0.2)$, breaking the implicit correlation between vorticity magnitude and geometry identity; (iv)~\emph{$\omega$-noise} adds Gaussian noise of std $0.05$ to the input field, providing local smoothing. These augmentations are designed to be compatible with the fixed-Chebyshev spatial operator (which is magnitude-agnostic by construction); applied to Transolver under matched settings, the same augmentations cause divergence (Appendix~\ref{sec:transolver_aug_appendix}).

\subsection{SepGRU: Traffic Forecasting (Learned Dynamics)}
\label{app:sepgru}

\paragraph{Task and protocol.}
We follow the DCRNN traffic forecasting protocol~\citep{li2018dcrnn}. Each model observes 12 historical 5-min speed values and predicts the next 12 steps. The cross-city transfer setting trains all learnable parameters on METR-LA and evaluates directly on PEMS-BAY. The reported main metric is the 30-min MAE on PEMS-BAY. Oracle gap is computed relative to a model trained and evaluated on PEMS-BAY under the same forecasting protocol.

\paragraph{Input features and zero-shot convention.}
At each sensor $i$ and time $t$, the dynamic input is
\begin{equation}
    \mathbf{x}_i^t = [\mathrm{speed}_i^t, \mathrm{tod}_i^t],
    \label{eq:traffic_input}
\end{equation}
where $\mathrm{tod}$ is the time-of-day encoding. SepGRU-v2 also uses local sensor statistics $\mathbf{f}_i \in \mathbb{R}^5$ to predict diffusion weights. These statistics are computed from the available unlabeled time series of the deployment city and contain no target training labels. Thus, the traffic experiment is a zero-label, cross-city transfer protocol: the target graph and simple target sensor summaries are available, but no PEMS-BAY supervised training is performed.

\paragraph{Step 1 (Shared temporal encoder).}
Each sensor history is encoded independently by a shared GRU:
\begin{equation}
    \mathbf{h}_i^{(\mathrm{enc})}
    =
    \mathrm{GRU}_{\mathrm{enc}}(\mathbf{x}_i^{1:T}).
    \label{eq:traffic_encoder}
\end{equation}
The same GRU weights are applied to every sensor in both cities. No sensor identity or graph-specific parameter enters this encoder.

\paragraph{Step 2 (Local-feature diffusion coefficients).}
SepGRU-v2 predicts node-wise multi-hop diffusion weights from local sensor statistics:
\begin{equation}
    \boldsymbol{\alpha}_i
    =
    \mathrm{softmax}(\mathrm{MLP}_{\mathrm{coeff}}(\mathbf{f}_i))
    \in \mathbb{R}^{K+1}.
    \label{eq:traffic_alpha}
\end{equation}
Here $\alpha_{i,0}$ weights the sensor's own hidden state and $\alpha_{i,k}$ weights the $k$-hop diffused hidden state. The coefficient MLP is shared across sensors and cities, so the node-wise adaptation is feature-conditioned rather than identity-conditioned.

\paragraph{Step 3 (Decoder with fixed graph diffusion).}
At each prediction step $\tau$, a shared GRUCell updates the temporal hidden state:
\begin{equation}
    \mathbf{h}^{(\tau)}
    =
    \mathrm{GRUCell}(\hat{\mathbf{y}}^{(\tau-1)}, \mathbf{h}^{(\tau-1)}).
    \label{eq:traffic_decoder_gru}
\end{equation}
Spatial information is then mixed using fixed powers of the normalized road adjacency $\mathbf{A}$:
\begin{equation}
    \mathbf{h}_i^{(\mathrm{out})}
    =
    \alpha_{i,0}\mathbf{h}_i^{(\tau)}
    +
    \sum_{k=1}^{K}
    \alpha_{i,k}(\mathbf{A}^k\mathbf{h}^{(\tau)})_i,
    \qquad
    \hat{y}_i^{(\tau)}
    =
    \mathbf{W}_{\mathrm{proj}}\mathbf{h}_i^{(\mathrm{out})}.
    \label{eq:traffic_diffusion_decoder}
\end{equation}
The adjacency matrix changes from METR-LA to PEMS-BAY, but it is never learned; it only appears through fixed matrix multiplications. All trainable weights---the GRU, coefficient MLP, and projection layer---are shared across graphs.

\paragraph{Baseline variants.}
SepGRU-v1 removes the learned coefficient predictor and uses fixed hop weights, testing whether node-wise adaptation is needed. NodeEmbedGRU adds a learnable embedding for each training sensor, testing the effect of source-node-specific parameters. EntangledGRU places graph diffusion inside the recurrent cell while keeping parameters shared, testing whether graph use inside temporal dynamics alone breaks transfer. The main result is that NodeEmbedGRU fails most severely, while EntangledGRU remains close to SepGRU, indicating that source-specific parameters are more damaging than fixed graph diffusion itself.

\section{GraphSAGE+GRU Stabilization Attempts}
\label{sec:sage_stabilization}

Dense GraphSAGE+GRU configurations were unstable or exceeded GPU memory on large networks. We therefore use a sparse-COO, node-budgeted implementation, trained for 150 epochs, which supports full-basin evaluation. As a shared-weight inductive model, its parameters remain meaningful on another basin.

\begin{table}[!htbp]
\centering
\caption{Controlled one-epoch Yangtze workload on Tesla V100S-PCIE-32GB: 1{,}388 subgraphs, 70{,}726 reach instances, and 1{,}200 hourly steps. Peak allocated GPU memory is reported. These are same-work timings, not matched-accuracy convergence times.}
\label{tab:sage_stabilization}
\small
\begin{tabular}{@{}lccc@{}}
\toprule
Model & Seconds/epoch & GPU GB & Parameters \\
\midrule
SepGFNO (FFT-IIR) & 72 & 2.5 & 27{,}924 \\
SepGFNO (sequential IIR) & 3{,}121 & 5.4 & 27{,}924 \\
GraphSAGE+GRU (sparse, ep150) & 396 & 24.2 & 67{,}651 \\
\bottomrule
\end{tabular}
\end{table}

The sequential control retains the same coefficient heads, five regimes, sparse upstream propagation, propagation rounds, and blending CNN as FFT-IIR; only the temporal recurrence is evaluated with a loop. Both SepGFNO solvers report one-epoch loss 0.1395.The sparse GraphSAGE+GRU comparison tests implementation-level efficiency under this workload; its cross-basin accuracy is in Table~\ref{tab:river}.

\section{SepTransolver: Adaptation and Multi-Geometry Stability}
\label{sec:transolver_aug_appendix}

Canonical Transolver~\citep{wu2024transolver} is a single-mesh, single-time-index method (forward pass: spatial coordinates + function values + scalar time index $\to$ field at one target time). It has no multi-geometry training mode. We construct \textbf{SepTransolver} by slotting Transolver's Physics-Attention spatial mixer into our space-time factorization protocol via five changes:
\begin{enumerate}[noitemsep,leftmargin=*]
\item \textbf{Time-windowed channel I/O}: scalar time index replaced by $T_w$-channel input/output ($\text{in\_dim} = n_\text{nf} + T_w$, $\text{out\_dim} = T_w$); predicts an entire window in one forward pass, making the model time-parallel (non-recurrent).
\item \textbf{Graph-free boundary features} replace absolute spatial coordinates ($x,y$), closing the Level-2 leak (\S\ref{sec:method}).
\item \textbf{Multi-mesh shared-weight batching} across heterogeneous meshes within a single batch (canonical Transolver processes one mesh per forward pass).
\item \textbf{Removal of scalar-$t$ conditioning} that would leak training-distribution time signature.
\item \textbf{Stability-tuned learning rate}: empirically required to be $\sim$30$\times$ smaller than canonical defaults to avoid divergence at multi-geometry training.
\end{enumerate}
We do not claim Physics-Attention is new; the contribution is the protocol that converts a single-geometry Physics-Attention solver into a multi-geometry-capable one.

\paragraph{Learning-rate sensitivity.}
At canonical-style LR ($1\!\times\!10^{-4}$), SepTransolver diverges to NaN within 30--90 epochs at $\geq 9$ training geometries (Table~\ref{tab:transolver_aug_full}, top rows). We swept LR $\in \{1\!\times\!10^{-4}, 3\!\times\!10^{-5}, 1\!\times\!10^{-5}\}$, gradient clipping $\in \{1.0, 5.0\}$, and augmentation on/off. LR$=3\!\times\!10^{-5}$ trains stably across all multi-geometry configurations. SepFNO is stable at LR$=1\!\times\!10^{-3}$ ($\sim$30$\times$ larger), reflecting the activation-scale sensitivity of softmax attention vs.\ polynomial mixing.

\paragraph{Architecture-tuned augmentation.}
Our SepFNO leak-control augmentation suite (rotation, boundary-feature dropout, $\omega$-scale, $\omega$-noise) is \emph{harmful} for SepTransolver in the original notch experiment. At 4g (LR$=3\!\times\!10^{-5}$), matched-aug yields $L_2{=}0.052$, no aug yields $L_2{=}0.046$ ($-12\%$). The $\omega$-scale/$\omega$-noise components are the primary culprits; intermediate (no-$\omega$) configuration yields $L_2{=}0.049$. We attribute this to the activation-scale sensitivity of Physics-Attention's softmax slice tokens. This notch sweep is distinct from the five-target, matched-augmentation sweep in Figure~\ref{fig:scaling}.

\begin{table}[h]
\centering
\caption{Triangle+notch SepTransolver configuration sweep, distinct from the five-target scaling table. \emph{Default LR} ($1\!\times\!10^{-4}$): diverges at $\geq 9$g. \emph{Tuned LR} ($3\!\times\!10^{-5}$): stable across these configurations.}
\label{tab:transolver_aug_full}
\small
\begin{tabular}{@{}lccccc@{}}
\toprule
Config & N & LR & Aug & Final $L_2$ & Status \\
\midrule
SepTransolver, default LR & 4  & 1e-4 & matched & 0.045 & stable \\
SepTransolver, default LR & 9  & 1e-4 & matched & --- & NaN ep 50 (best 0.038) \\
SepTransolver, default LR & 15 & 1e-4 & matched & --- & NaN ep 50 (best 0.056) \\
\midrule
SepTransolver, tuned LR & 4 & 3e-5 & matched & 0.052 & stable \\
SepTransolver, tuned LR & 4 & 3e-5 & none    & \textbf{0.046} & stable \\
SepTransolver, tuned LR & 9 & 3e-5 & matched & 0.034 & stable \\
SepTransolver, tuned LR & 9 & 3e-5 & none    & \textbf{0.022} & stable \\
SepTransolver, tuned LR & 15 (bigger) & 3e-5 & matched & 0.031 & stable \\
SepTransolver, tuned LR & 15 (bigger) & 3e-5 & none    & \textbf{0.021} & stable \\
\midrule
SepFNO base                & 9  & 1e-3 & matched & 0.031 & stable \\
SepFNO base                & 15 & 1e-3 & matched & 0.030 & stable \\
SepFNO bigger              & 15 & 1e-3 & matched & 0.028 & stable \\
\bottomrule
\end{tabular}
\end{table}

\paragraph{Mechanism we hypothesize but do not prove.}
Physics-Attention uses a softmax over slice tokens weighted by a learned temperature. Multi-geometry training varies the magnitude scale across batches; the softmax is sensitive to this, leading either to peaked attention (vanishing gradients on off-peak slices) or large activations (overflow). Smaller LR shrinks per-step updates enough to avoid the unstable regime, and removing $\omega$-scale aug eliminates an extra source of magnitude perturbation. SepFNO's spatial operator is a fixed polynomial of the adjacency---no softmax, no learned temperature---so magnitude variation scales its outputs linearly without instability.

\section{Related Cross-Geometry PDE Benchmarks and How Our Setup Differs}
\label{sec:related_benchmarks}

Operator learning on irregular geometries is an active area, and several benchmarks evaluate generalization across mesh, geometry, or parameter shifts. Here we survey the most relevant efforts and indicate where each differs from the multi-geometry knowledge-accumulation regime evaluated in \S\ref{sec:pde}.

\begin{table}[h]
\centering
\caption{Related cross-geometry / mesh-generalization benchmarks. None tests how cross-geometry error scales with the number and diversity of training geometries, the regime our multi-target scaling experiment targets.}
\label{tab:related_benchmarks}
\small
\setlength{\tabcolsep}{4pt}
\begin{tabular}{@{}p{3.1cm}p{3.5cm}p{6.8cm}@{}}
\toprule
Benchmark / paper & Scope & How it differs from our regime \\
\midrule
PDEBench~\citep{takamoto2022pdebench} & Broad SciML benchmark for time-dependent PDEs (Burgers, Navier--Stokes, etc.) with standardized data, baselines, metrics & Tests model performance on each PDE family separately; not designed for joint training across many distinct geometries with zero-shot transfer to a held-out geometry. \\
\addlinespace[2pt]
SURF~\citep{kunzli2024surf} & GNN fluid simulator generalization across mesh resolution, parameter ranges, topology shifts within engineering systems (valves, turbomachinery) & Within-system OOD generalization rather than cross-distinct-shape knowledge accumulation; does not run the ``train on $N$ shapes, ask if more $N$ helps'' protocol that motivates our scaling figure. \\
\addlinespace[2pt]
\citet{schmocker2024meshgraphnets} & MGN trained on cylinder-flow with various obstacle shapes and multi-object configs & Cross-shape transfer within a single physical setup (cylinder flow); the shape variation is over obstacle inserts, not fundamentally different mesh topologies (T-beam, hourglass, star, etc.) as in our setup. \\
\addlinespace[2pt]
Geo-FNO~\citep{li2023geofno} & FFT operators on irregular geometries via learned deformations to latent regular grids & A method/architecture, not a standardized benchmark; the deformation is learned per-shape and time is advanced autoregressively. \\
\addlinespace[2pt]
GINO~\citep{li2023gino} & Geometry-aware neural operator for large-scale 3D PDEs & A method paper with bespoke geometry-aware encoders; spatial operators carry mesh-dependent learned weights (different from our graph-free constraint). \\
\addlinespace[2pt]
Transolver~\citep{wu2024transolver} & Physics-Attention slice mechanism for irregular meshes & A method paper; published evaluations are within a single mesh distribution per task. We adapt to our cross-geometry windowed protocol as \emph{SepTransolver} (\S\ref{sec:transolver_aug_appendix}). \\
\bottomrule
\end{tabular}
\end{table}

\paragraph{Position relative to existing work.}
The broad area of geometry generalization in PDE learning is well established. What we are not aware of, and what we therefore construct in \S\ref{sec:pde}, is a controlled setup that specifically tests \emph{cross-geometry knowledge accumulation}: train jointly on many distinct mesh topologies, evaluate zero-shot on a held-out geometry, and measure how performance scales with the number and diversity of training geometries. This is the scaling-with-$N$-training-geometries axis (Fig.~\ref{fig:scaling}), and it is the axis where our principle's value becomes empirically visible.

\paragraph{On the fairness of baseline adaptations.}
We do not claim that the architectures evaluated in Table~\ref{tab:pde_baselines} are intrinsically incapable of cross-geometry transfer. Their published validation regimes — single physical systems with mesh adaptation, OOD shifts within an engineering family, obstacle-shape variation within cylinder flow — differ from ours, and our results reflect performance under the specific adaptations we made (e.g., adding boundary-feature inputs, channel-style time windows, multi-mesh batching). For Transolver specifically, the adaptation is structural (the original takes a scalar time index and predicts a single field; we extend this to time-windowed channel I/O), and we report the adaptation in detail under the name \emph{SepTransolver}. Readers who can identify suboptimalities in our adaptations will materially improve the comparison; every adapted launch script and the model code are available at \url{https://github.com/chaopengshen/gsno_crossgraph}.

\section{Citation Network Experiments}
\label{sec:citation_appendix}

We test the separation principle on citation network node classification using the ArnetMiner dataset (GDABench~\citep{gdabench2024}): DBLP, ACM, and Citation graphs with 6{,}775-dimensional bag-of-words text features and multi-class labels (5 research areas).

\paragraph{SepHybrid architecture.}
SIGN-style~\citep{frasca2020sign} pre-smoothing ($\mathbf{x}_\text{aug} = [\mathbf{x}, \mathbf{A}\mathbf{x}, \mathbf{A}^2\mathbf{x}, \ldots, \mathbf{A}^K\mathbf{x}]$) provides multi-scale graph features. An MLP processes the concatenated input, followed by 3 iterative refinement layers: $\mathbf{h} \leftarrow \mathbf{h} + \text{MLP}_k([\mathbf{h}; \mathbf{A}\mathbf{h}])$. All learned parameters are graph-free; the graph enters only through fixed matrix powers during pre-smoothing and iterative steps.

\begin{table}[h]
\centering
\caption{Citation cross-graph classification (Macro-F1 \%). All 6 source$\to$target pairs.}
\label{tab:citation}
\small
\begin{tabular}{@{}lccccccc@{}}
\toprule
Model & D$\to$A & D$\to$C & A$\to$D & A$\to$C & C$\to$D & C$\to$A & Avg \\
\midrule
SepHybrid & \textbf{59.1} & \textbf{65.8} & \textbf{61.7} & \textbf{68.9} & 67.3 & 66.0 & \textbf{64.8} \\
GCN-3L & 57.0 & 65.3 & 58.5 & 66.9 & \textbf{67.9} & \textbf{66.7} & 63.7 \\
GCN-2L & 56.3 & 64.0 & 59.8 & 67.0 & 66.9 & 66.0 & 63.3 \\
MLP & 41.6 & 45.6 & 47.0 & 47.2 & 50.2 & 45.2 & 46.1 \\
\bottomrule
\end{tabular}
\end{table}

SepHybrid (64.8\%) outperforms GCN-3L (63.7\%) by +1.1 points, winning 4 of 6 transfer pairs. The result is modest compared to the spatiotemporal domains, consistent with the observation that static node classification with homophilic graphs provides less room for improvement through separation.

\section{NSE Geometries and Data}
\label{sec:nse_data}

Table~\ref{tab:scaling_new} varies the number of \emph{training} NSE geometries: 5/6/7/9/12 for SepFNO and 6/7/9/15 for SepTransolver. Its held-out evaluation set is the same five newly generated car, cup, spiral, pistol, and saw geometries for every row; none appears in a training pool, and there is no target fine-tuning. Their meshes have 1{,}803, 981, 3{,}817, 786, and 1{,}930 nodes. Data are generated with viscosity 0.01, quasi-periodic forcing amplitude 5, 50 samples, $t_{\rm end}=9$, 1{,}000 solver steps, and every-fourth-step saving ($T=251$). SepFNO results use seeds 101, 202, and 303. SepTransolver is a separate single-run companion under matched augmentation. The SepFNO uncertainty column is seed SD, not across-target SD.

\begin{table}[!htbp]
\centering
\caption{Varying-size NSE training pools, evaluated zero-shot on the same five held-out car/cup/spiral/pistol/saw geometries (relative vorticity $L_2$). SepFNO target cells are three-seed means; SepTransolver is single-run.}
\label{tab:scaling_new}
\small
\setlength{\tabcolsep}{4pt}
\begin{tabular}{@{}llcccccc@{}}
\toprule
Model & Pool & Car & Cup & Spiral & Pistol & Saw & Mean $\pm$ seed SD \\
\midrule
SepFNO & 5g & .0652 & .1109 & .1668 & .0563 & .0623 & $.0923\pm.0037$ \\
 & 6g & .0717 & .1147 & .1759 & .0614 & .0675 & $.0982\pm.0065$ \\
 & 7g & .0591 & .1073 & .1544 & .0511 & .0549 & $.0854\pm.0068$ \\
 & 9g & .0426 & .0997 & .1177 & .0394 & .0360 & $.0671\pm.0014$ \\
 & 12g & .0431 & .0969 & .1176 & .0379 & .0359 & $.0663\pm.0005$ \\
\midrule
SepTransolver & 6g & .0490 & .1352 & .1938 & .0817 & .0468 & .1013 \\
 & 7g & .0495 & .1298 & .1754 & .0763 & .0473 & .0957 \\
 & 9g & .0316 & .1201 & .1195 & .0489 & .0268 & .0694 \\
 & 15g & .0285 & .1087 & .1174 & .0467 & .0248 & .0652 \\
\bottomrule
\end{tabular}
\end{table}

At fixed SepTransolver pool size, adding sinusoidal channel to a 4g base changes five-target mean error from 0.0994 to 0.0754, adding star to 0.0827, and adding hourglass to 0.1102. Thus composition may help or hurt. Target-only oracle errors under the same data-generation settings are 0.0175 (car), 0.0314 (cup), 0.0326 (spiral), 0.0135 (pistol), and 0.0171 (saw); best reported cross-geometry/oracle ratios span 1.10--3.54. These oracle comparisons do not establish a universal near-parity result.

Table~\ref{tab:pde_six} tests a different protocol: fixed checkpoints trained on the same four base geometries (ellipse with triangle hole, mesh rectangle with triangle hole, rectangle with triangle and circular holes, and six-petal flower with circular hole) are evaluated without adaptation on all six other geometries in a separate ten-geometry pool. These six held-out meshes are not the five car/cup/spiral/pistol/saw targets of Table~\ref{tab:scaling_new}. Vector velocity is reconstructed for predictions and targets from $\nabla^2\psi=-\omega$ with zero boundary streamfunction, $u=\partial_y\psi$, and $v=-\partial_x\psi$. Relative velocity $L_2$ combines nodes, output times, and both components per window, then averages over windows. Pressure is eliminated and neither predicted nor stored.

\begin{table}[!htbp]
\centering
\caption{Fixed four-geometry NSE training set; zero-shot evaluation on its six held-out geometries from the ten-geometry pool. Each cell is relative vorticity $L_2$ / reconstructed velocity $L_2$. This test set differs from Table~\ref{tab:scaling_new}.}
\label{tab:pde_six}
\footnotesize
\setlength{\tabcolsep}{3pt}
\begin{tabular}{@{}lccccc@{}}
\toprule
Target & SepFNO & MLP\_FNO & GCN\_FNO & GraphSAGE\_FNO & MeshGraphNet \\
\midrule
L-shape & .0478/.0258 & .1356/.0768 & .2964/.1859 & .0813/.0439 & .2479/.1488 \\
Crescent & .0324/.0192 & .0885/.0576 & .2201/.1774 & .0441/.0237 & .1894/.1557 \\
Hourglass & .0539/.0291 & .1547/.0882 & .2887/.1794 & .0971/.0535 & .2433/.1387 \\
Sinusoidal & .0563/.0400 & .0901/.0684 & .2669/.2245 & .0581/.0412 & .2283/.1929 \\
Star & .0586/.0323 & .1113/.0740 & .2653/.2005 & .0643/.0394 & .2329/.1725 \\
Notch & .0403/.0237 & .1026/.0700 & .2759/.1962 & .0593/.0376 & .2397/.1681 \\
\midrule
Mean & .0482/.0283 & .1138/.0725 & .2689/.1940 & .0674/.0398 & .2303/.1628 \\
\bottomrule
\end{tabular}
\end{table}

On the external Schmöcker unsteady cylinder-flow benchmark~\citep{schmocker2024meshgraphnets}, models train on \texttt{standard\_cylinder} and evaluate on that geometry and zero-shot on \texttt{cylinder\_stretch}. The comparison uses 32 simulations evaluated by both models (11 with Reynolds number below 47). Errors are raw $v_x$ RMSE over 20 steps. The native MeshGraphNet checkpoint was at epoch 20 and undertrained (one-step error $11.7\times10^{-3}$ versus the published $2.32\times10^{-3}$); therefore this comparison does not establish superiority over a converged baseline. High-Re SepFNO errors reflect phase drift rather than numerical NaNs.

\begin{table}[!htbp]
\centering
\caption{External benchmark on 32 matched simulations: $v_x$ RMSE $\times10^{-3}$ as median / mean. Standard is in distribution; stretch is zero-shot geometry transfer.}
\label{tab:schmocker}
\small
\setlength{\tabcolsep}{4pt}
\begin{tabular}{@{}lccccc@{}}
\toprule
Re bin & $n$ & MGN standard & SepFNO standard & MGN stretch & SepFNO stretch \\
\midrule
$<47$ & 11 & 121.1 / 114.7 & 53.2 / 52.3 & 163.0 / 163.3 & 76.0 / 143.4 \\
47--100 & 8 & 136.0 / 111.1 & 142.5 / 195.3 & 147.4 / 165.1 & 73.7 / 235.1 \\
$\geq100$ & 13 & 145.1 / 124.5 & 128.8 / 222.1 & 160.0 / 178.9 & 376.4 / 303.6 \\
All & 32 & 136.5 / 117.8 & 53.6 / 157.0 & 159.5 / 170.1 & 125.7 / 231.4 \\
\bottomrule
\end{tabular}
\end{table}

For the 4-geometry training setting used in Table~\ref{tab:pde_baselines}, the training meshes are \emph{Ellipse with Triangle Hole}, \emph{Mesh Rectangle with Triangle Hole}, \emph{Rectangle with Triangle and Circular Holes}, and \emph{Six-Petal Flower with Circular Hole}. The held-out zero-shot test mesh is \emph{Triangular with a Notch}. 

\begin{figure}[!htbp]
\centering
\includegraphics[width=\textwidth]{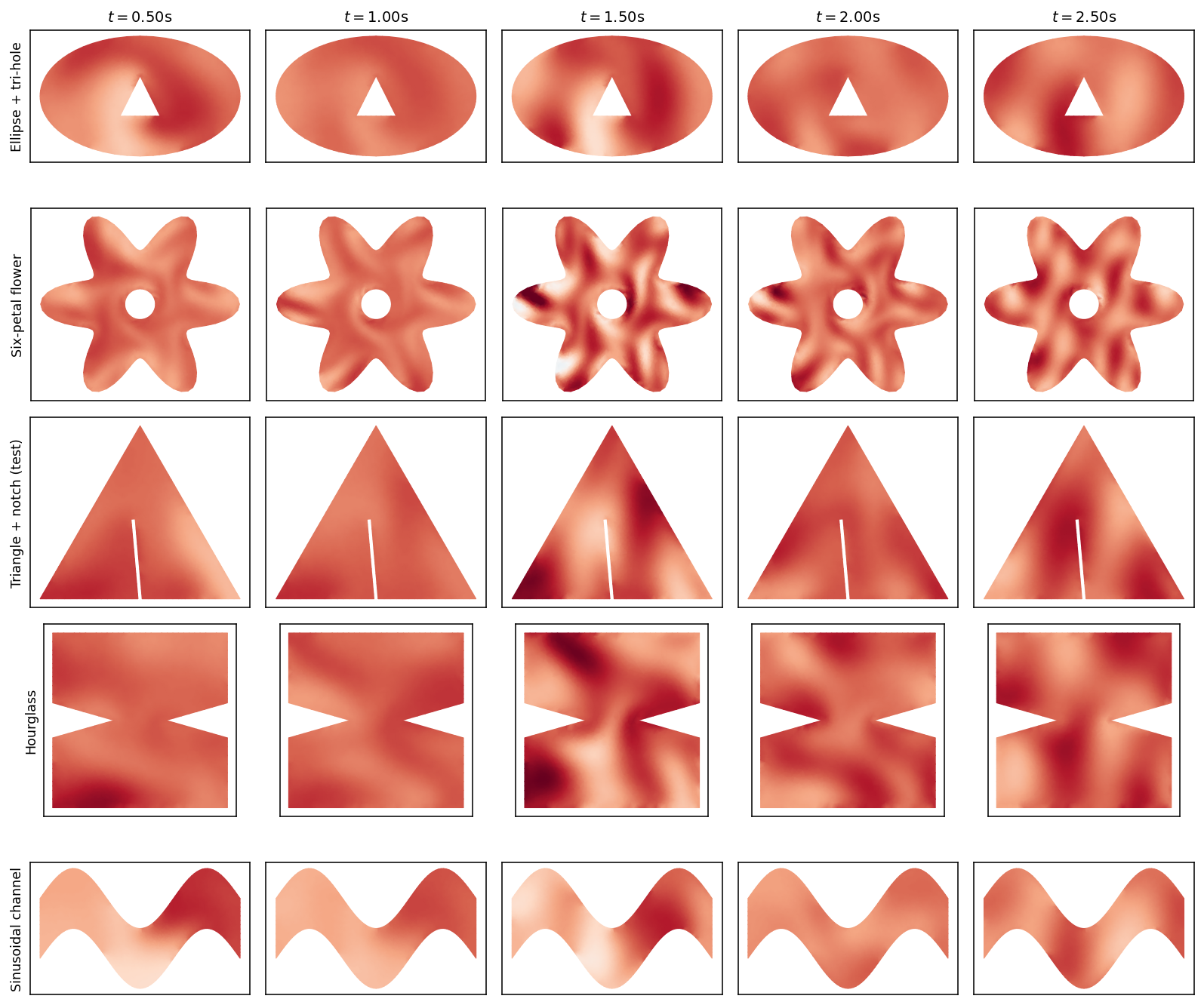}
\caption{NSE vorticity fields on five training/test geometries (ellipse+triangle hole, six-petal flower, triangle+notch test, hourglass, sinusoidal channel) at five time instants. Quasi-periodic forcing drives complex spatiotemporal dynamics on each irregular mesh. The triangle+notch (third row) is the held-out zero-shot test geometry.}
\label{fig:nse_geometries}
\end{figure}

\begin{figure}[!htbp]
\centering
\includegraphics[width=0.95\textwidth]{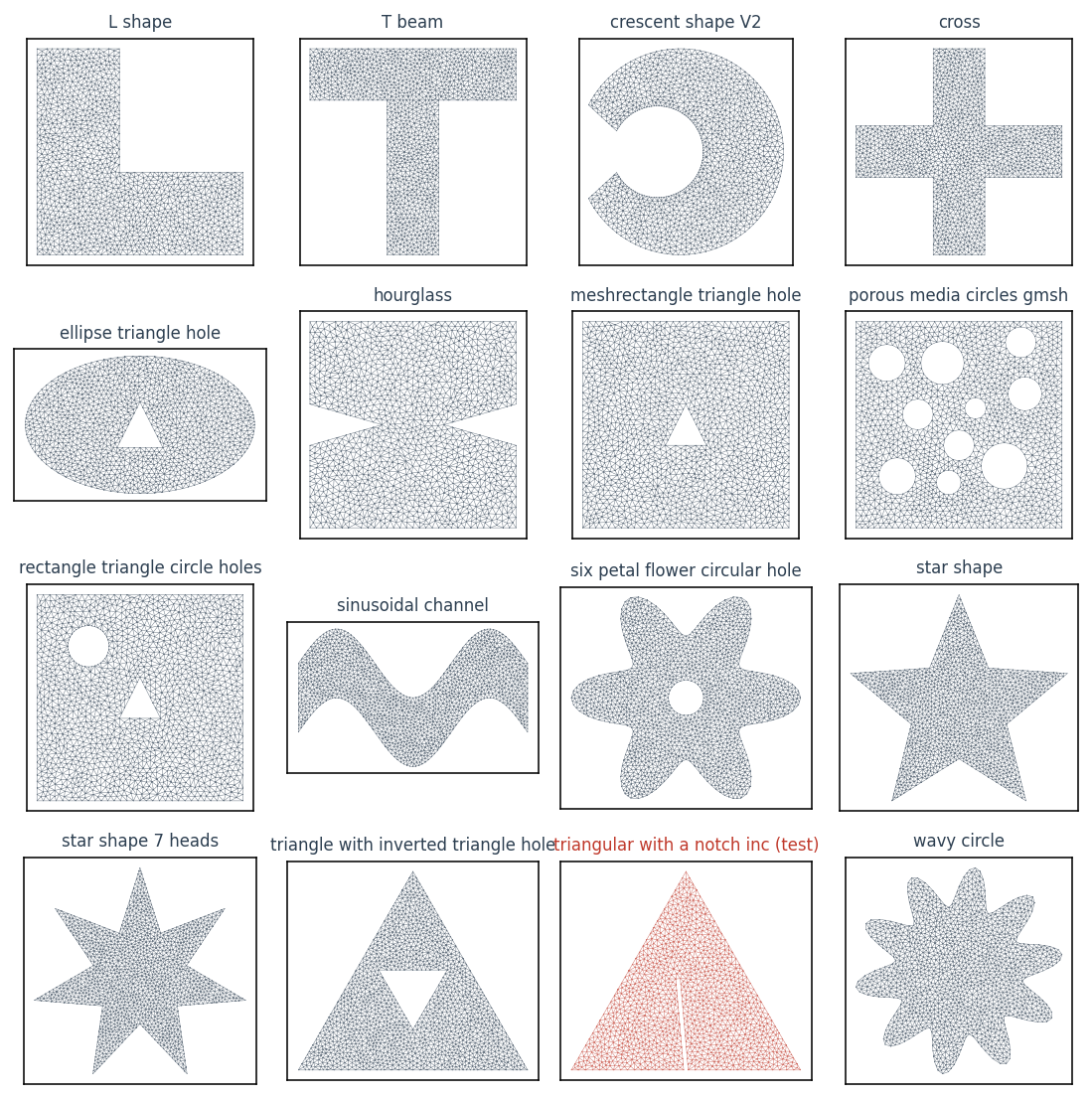}
\caption{All 16 NSE meshes used in the cross-geometry experiments: 15 training geometries (black mesh) and the held-out test geometry triangle+notch (red, bottom row). Mesh size varies $\sim$1{,}200--1{,}700 nodes; geometries include simple convex shapes (ellipse, hexagon), shapes with internal obstacles (multi-holes, porous media), and shapes with sharp boundary features (hourglass, star, crescent, sinusoidal channel).}
\label{fig:nse_gallery}
\end{figure}

Figure~\ref{fig:nse_geometries} shows representative vorticity rollouts on five of the training/test geometries; the full gallery of all 15 training meshes plus the held-out test mesh is shown in Figure~\ref{fig:nse_gallery}. We use 2D incompressible Navier--Stokes on irregular Delaunay-triangulated domains ($\sim$1{,}200--1{,}700 nodes per mesh) with interior obstacles. A quasi-periodic body force ($\text{amp}=5$, $\nu=0.01$) drives complex spatiotemporal dynamics on each mesh, generating multi-step vorticity rollouts; the cross-geometry experiments train on $N$ training geometries and hold out triangle+notch for zero-shot evaluation.

\section{Code Availability}
\label{sec:code_availability}

Code is available at \url{https://github.com/chaopengshen/gsno_crossgraph}.



\end{document}